%% file: main.tex
\documentclass[11pt]{article}
\usepackage[preprint]{acl} % "review" & "final" & "preprint"

\input{Sec/preamble}

\title{\textit{\gradientRGB{VLN-MME}{128,0,128}{210, 153, 153}
}: Diagnosing MLLMs as Language-guided \\ Visual Navigation agents}

\author{Xunyi Zhao\thanks{Equal Contribution, $\dagger$ Project lead}, Gengze Zhou$^{*\dagger}$, Qi Wu \vspace{5pt}\\
Australian Institute for Machine Learning, Adelaide University \\
\vspace{5pt}
\texttt{\{xunyi.zhao, gengze.zhou, qi.wu01\}@adelaide.edu.au} \vspace{2pt}\\
\href{https://github.com/billzhao1030/VLN-MME}{\color{purple} https://github.com/billzhao1030/VLN-MME}
}

\begin{document}
\maketitle

\input{Sec/abstract}
\input{Sec/introduction}
\input{Sec/relatedwork}
\input{Sec/method}
\input{Sec/experiment}
\input{Sec/conclusion}

\clearpage
\input{Sec/limitation}
\bibliography{custom}

\input{Sec/appendix}

\end{document}

%% file: Sec/preamble.tex
\usepackage{times}
\usepackage{latexsym}
\usepackage[T1]{fontenc}
\usepackage[utf8]{inputenc}
\usepackage{microtype}
\usepackage{inconsolata}

\usepackage{graphicx}
\usepackage{hyperref}
\usepackage{url}
\usepackage{gradient-text}
\usepackage{graphicx}
\usepackage{booktabs}      
\usepackage{multirow}
\usepackage[dvipsnames]{xcolor}
\usepackage{colortbl}
\usepackage{amssymb}
\usepackage{amsmath}
\usepackage{float}
\usepackage{pifont}

\def\ours{VLN-MME}

%% file: Sec/abstract.tex
\begin{abstract}
\label{sec:abstract}
Multimodal Large Language Models (MLLMs) have demonstrated remarkable capabilities across a wide range of vision-language tasks. However, their performance as embodied agents, which requires multi-round interaction with spatial reasoning and sequential action prediction, needs further exploration. Our work investigates this potential in the context of Vision-and-Language Navigation (VLN) by introducing a unified and extensible simulation-free evaluation framework to probe MLLMs as zero-shot agents, named \ours. Simplifying the evaluation with a highly modular and accessible design streamlines experiments, enabling structured comparisons and component-level ablations across diverse MLLM architectures, agent designs, and navigation tasks. Crucially, enabled by \ours, we observe that enhancing prevalent agents with Chain-of-Thought (CoT) reasoning and self-reflection leads to an unexpected performance decrease. This suggests MLLMs exhibit poor context awareness in embodied navigation tasks; although they can follow instructions and structure their output, their 3D spatial reasoning fidelity is low. Furthermore, we demonstrate that agent performance could be largely improved with simple failure cases in context learning. \ours\ lays the groundwork for systematic evaluation of general-purpose MLLMs in embodied navigation settings and reveals limitations in their sequential decision-making capabilities. We believe these findings offer crucial guidance for MLLM post-training as embodied agents. 
\end{abstract}

%% file: Sec/introduction.tex
\section{Introduction}
\label{sec:introduction}

The rapid advancement of Multimodal Large Language Models (MLLMs) has catalyzed a shift from static vision-language tasks toward the development of embodied agents capable of dynamic, interactive decision-making. Central to this transition is the evaluation of 3D spatial reasoning. However, prior benchmarks have largely focused on static spatial Question Answering or simple navigation or manipulation tasks within low-fidelity, synthetic environments~\citep{yang2025thinking, daxberger2025mm, yang2025embodiedbench, cheng2025embodiedeval}. Crucially, these settings lack substantial changes in 3D environmental states and do not require the maintenance of long-term spatial memory.
% The rapid advancement of Multimodal Large Language Models (MLLMs) has raised interest in deploying them as embodied agents, moving beyond static vision-language tasks to dynamic, interactive decision-making. 
% In this context, Vision-and-Language Navigation (VLN)~\citep{anderson2018r2r} emerges as a crucial and challenging paradigm to evaluate the MLLM's 3D reasoning ability. Successfully navigating a 3D environment based on instructions requires more than pattern recognition; it fundamentally tests an agent's spatial understanding, its ability to plan and foresee the consequences of its actions, and its use of long-term memory to ground an extended plan. When navigation involves multi-round dialogue, it further probes the model's capacity for contextual reasoning. However, despite VLN's potential as a comprehensive benchmark for these core agentic skills, progress in systematically evaluating MLLMs is constrained by the limitations of existing evaluation pipelines.
In contrast, Vision-and-Language Navigation (VLN)~\citep{anderson2018r2r} emerges as a far more challenging paradigm for evaluating these abilities. Successfully navigating a 3D environment requires more than simple pattern recognition; it fundamentally tests an agent's spatial understanding, its ability to maintain awareness, foresee the consequences of its actions across dynamic viewpoint changes, and the use of long-term memory to ground extended plans. Furthermore, when navigation involves multi-round interaction, it probes the model's capacity for contextual reasoning. However, despite VLN's potential as a comprehensive benchmark for these core agentic skills, progress in systematically evaluating MLLMs is constrained by the limitations of existing evaluation pipelines.

% First, embodied navigation tasks typically run in high-fidelity simulators such as Matterport3D~\citep{chang2017matterport3d} or Habitat~\citep{savva2019habitat}. The evaluation cost grows sharply when large models are deployed as VLN agents in multi-round settings that require frequent interaction with the environment. Second, the existing VLN benchmarks are diverse~\citep{anderson2018r2r, qi2020reverie, anderson2020rxr}, and a single dataset can contain thousands of navigation trajectories, making comprehensive evaluation with large MLLM agents a prohibitively time-consuming and computationally heavy process. Third, prior studies often focus on improving success metrics with different LLMs, and rarely offer principled error analyses, which limits comparability and obscures the true contributions of model capability versus agent design.

First, embodied navigation typically relies on high-fidelity simulators such as Matterport3D~\citep{chang2017matterport3d} or Habitat~\citep{savva2019habitat}. When deploying large MLLMs in these multi-round, interactive settings, the computational cost grows sharply. This issue is compounded by the high volume of trajectories across diverse benchmarks~\citep{anderson2018r2r, qi2020reverie, anderson2020rxr}, making comprehensive testing prohibitively time-consuming. As a result, the computational burden of exhaustive evaluation has pushed prior studies toward largely metric-driven approaches, prioritizing end-to-end success rates over diagnostic clarity. This lack of principled error analysis obscures the underlying model behavior, making it difficult to assess critical capabilities such as generalization, robustness, or the specific alignment between visual perception and instruction following.

% More critically, recent approaches to evaluating MLLMs in VLN have gaps in understanding model behavior. On one hand, some works utilize end-to-end success metrics alone and are insufficient for understanding agent behavior.
% On the other hand, dedicated evaluation suites like NavBench~\citep{qiao2025navbench}, while comparing different models and tasks, do not systematically consider the crucial impact of varying agent designs. Consequently, the community still lacks a deeper understanding of how these models perform. Specifically, there is minimal fine-grained analysis of success and failure cases, error types, or patterns in agent decision-making. To address these limitations, we developed our own modular evaluation framework, designed specifically to diagnose MLLM behavior in navigation tasks. The necessity for such a framework is highlighted by a comparison with existing benchmarks in Table~\ref{tab:benchmark_comparison}. Without the kind of diagnostic insights our approach provides, it is difficult to assess generalization, robustness, or the alignment between visual perception and instruction-following capabilities in MLLMs. As a result, progress in the field remains largely metric-driven, with little clarity on the underlying model behavior.

More critically, recent evaluation suites like NavBench~\citep{qiao2025navbench} have attempted to standardize this process and successfully unify the evaluation of different tasks and models. However, it is restricted to a single agent design and not systematically varying agent designs; it becomes impossible to decouple the intrinsic capabilities of the MLLM from the efficacy of specific prompting or planning strategies. As a result, the community lacks a fine-grained understanding of whether failures stem from the model's reasoning limitations or suboptimal agent engineering.

% \input{Tab/benchmarks}

% In response to these gaps, we propose the \textbf{Vision Language Navigation Multi-Model Evaluation (VLN-MME)}, a novel framework designed to address these limitations. Our approach is built on a modular and simulator-free architecture that prioritizes accessibility and reproducibility. Crucially, instead of focusing on high-level success metrics, we contribute a detailed error analysis that breaks down agent performance to evaluate core capabilities. This allows for a deeper understanding of an MLLM's proficiency in instruction following, spatial understanding, and historical sequential reasoning for long-horizon tasks.

In response to these challenges, we propose the \textbf{Vision Language Navigation Multi-Model Evaluation (VLN-MME)}, a novel framework designed to address these limitations. Our approach features a modular and simulator-free architecture that preserves navigational semantics while eliminating the setup complexity and computational overhead of traditional pipelines. Crucially, we move beyond high-level success metrics to provide a detailed error analysis, dissecting agent performance to diagnose core proficiencies in instruction following, spatial understanding, and historical reasoning.

% Our contributions could be summarized as:
% \begin{itemize}
%     \item We present a unified evaluation framework that enables structured, comparable assessment of different MLLMs, agents, and VLN tasks under a consistent interface.
%     \item We introduce a simulator-free design that preserves navigational semantics while significantly reducing setup complexity and enabling broader accessibility.
%     \item We curate and publish VLN data, environments, and configuration artifacts on public platforms to streamline benchmarking and reproducibility.
%     \item We conduct an extensive and insightful error analysis that uncovers behavioral patterns and limitations in MLLMs' navigation reasoning.
% \end{itemize}

Our contributions are summarized as follows: 
\begin{itemize} 
    \item We introduce VLN-MME, a unified and modular framework that systematically evaluates the interplay between Model, Agent, and Task, addressing the limitation of fixed-agent designs in prior benchmarks. \item We design a simulator-free evaluation pipeline that preserves essential navigational semantics while drastically reducing computational overhead and setup complexity to enhance accessibility. \item We provide a comprehensive diagnostic analysis that moves beyond success metrics, categorizing fine-grained failure modes to uncover specific deficiencies in MLLM spatial reasoning and instruction following. 
    \item We release a standardized suite of processed datasets and environmental artifacts to facilitate reproducible research and streamline the benchmarking process. 
\end{itemize}

% This work aims to establish a standardized foundation for studying MLLMs in embodied environments, pushing the field beyond leaderboard metrics toward a deeper understanding of model behavior.

% This work establishes a rigorous foundation for studying MLLMs in embodied environments, shifting the focus from leaderboard rankings to a deeper, mechanistic understanding of model behavior.

%% file: Sec/relatedwork.tex
\section{Related Works}
\label{sec:related_works}

\paragraph{Evaluating MLLMs in Spatial and Embodied Contexts} Comprehensive benchmarks have emerged to test a wide spectrum of MLLM abilities~\citep{chaoyou2023mme, liu2024mmbench, li2024mvbench, yue2024mmmu, yu2024mm, lu2023mathvista, fei2025path}, ranging from basic perception to complex cognition. Within this landscape, specific efforts have focused on assessing 3D spatial reasoning~\citep{yang2025thinking, daxberger2025mm, xu2025spatialbench, liao2024reasoning, li2024behavior}. However, the majority of these benchmarks rely on static QA formats, where the model provides a single response to a fixed visual input, but not for continuous state tracking.
To assess sequential reasoning, several benchmarks focus on long-horizon tasks; however, these are largely restricted to digital domains like web browsing and application usage~\citep{deng2023mind2web, trivedi2024appworld, tao2025mmsearch, wang2025odysseybench}. More recent embodied benchmarks attempt to bridge this gap but face distinct limitations. Some works, such as 3DMEM-Bench~\citep{hu20253dllm}, focus primarily on high-level planning. While effective for evaluating abstract reasoning, these approaches often overlook the fine-grained environmental interactions required for realistic agent execution. Conversely, benchmarks like~\citep{yang2025embodiedbench, cheng2025embodiedeval} operate within low-fidelity, synthetic environments. Although they incorporate manipulation tasks, these settings generally lack photorealistic visual complexity and do not involve the massive 3D environmental state changes inherent to large-scale navigation. Consequently, these tasks rarely demand the long-term spatial memory or the rigorous sequential reasoning required to operate in dynamic, photorealistic spaces.

\paragraph{MLLMs for Vision-and-Language Navigation} The integration of MLLMs into robotics has inspired new paradigms for VLN. Early efforts leveraged LLMs as copilots to guide specialist agents~\citep{qiao2023march}, while recent works employ off-the-shelf MLLMs as zero-shot agents through elaborate prompting~\citep{zhou2024navgpt, long2023discuss, chen2024mapgpt, dong2025se}. Concurrently, other studies have explored fine-tuning MLLMs directly on navigation data~\citep{zhou2025navgpt2,lin2024navcot,pan2023langnav, zheng2023towards} or adapting pre-trained video models~\citep{zhang2024navid, zhang2024uni, cheng2024navila}.
Despite this progress, evaluation remains fragmented and costly due to reliance on simulators. While recent efforts like SAME~\citep{zhou2025same} and NavBench~\citep{qiao2025navbench} attempt to standardize task evaluation, they face distinct limitations: the former does not specifically target MLLM agents, while the latter is restricted to fixed agent design. Consequently, existing frameworks often focus on aggregate metrics, lacking the fine-grained diagnostics required to understand specific failure modes. We bridge this gap with a unified, simulator-free framework that jointly evaluates MLLMs, agent designs, and diverse tasks.

%% file: Sec/method.tex
\section{Method}
\label{sec:method}

\subsection{Modular VLN Evaluation Framework}
\label{sec:framework-design}

% To enable systematic and reproducible research on MLLMs in embodied settings, we designed and implemented a modular software stack for VLN evaluation. Our architecture enforces a clean separation of concerns between its primary components: the model, the agent, and the environment. This modularity empowers us to seamlessly interchange different MLLMs, implement novel agent designs, or introduce new datasets for structured comparisons and component-level ablations. The high-level architecture of our framework is illustrated in Figure~\ref{fig:framework-diagram}.

\begin{figure}[h]
    \centering
    \includegraphics[width=\columnwidth]{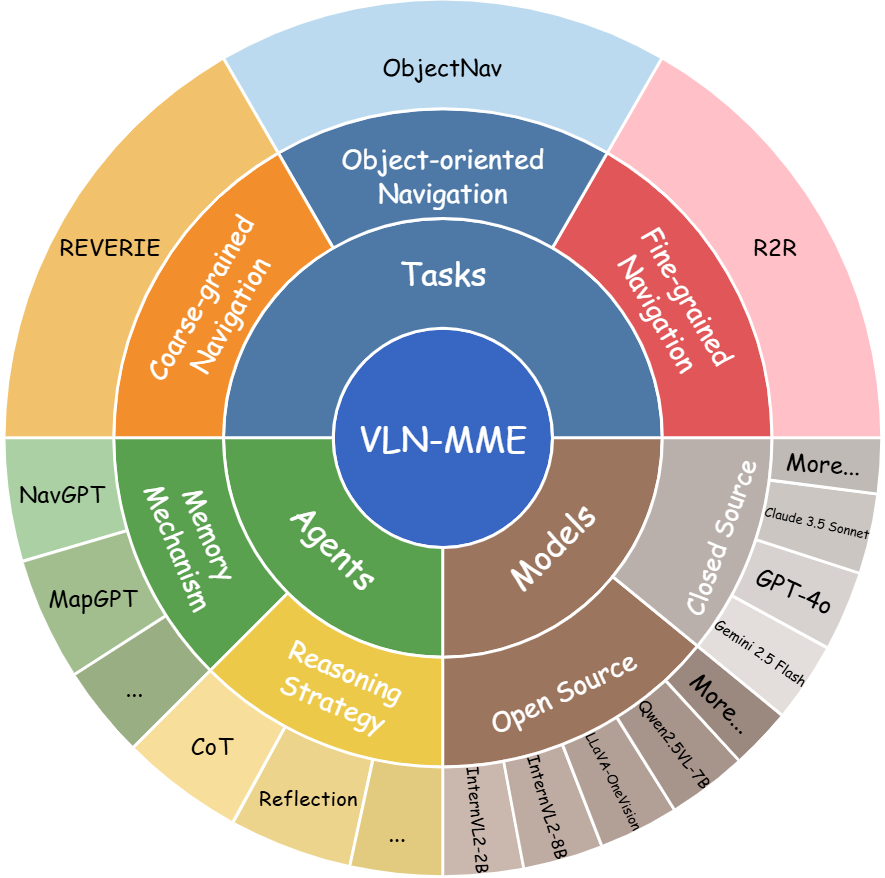}
    \caption{The composition of the VLN-MME.}
     % benchmark, detailing the diverse set of \textbf{Tasks}, \textbf{Agents}, and \textbf{Models} it supports
    \label{fig:vlnmme}
\end{figure}

To enable systematic assessment of MLLMs' spatial reasoning, long-horizon planning, and sequential decision-making capabilities in embodied settings, we design a modular software stack that cleanly separates three core components: Model, Agent, and Task (Figure~\ref{fig:framework-diagram} \&~\ref{fig:vlnmme}). This modularity empowers us to seamlessly interchange each component independently, enabling structured comparisons and component-level ablations to isolate the source of success or failure. Our framework is built upon three primary abstractions:

\begin{figure}[h]
    \centering
    \includegraphics[width=\columnwidth]{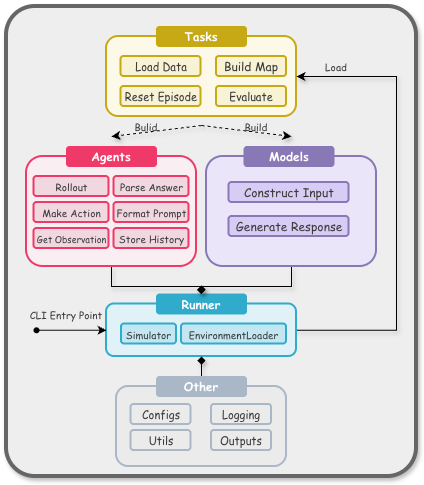}
    \caption{A high-level structure for the benchmark, centered on the interplay between \textbf{Tasks}, \textbf{Agents}, and \textbf{Models}.}
    \label{fig:framework-diagram}
\end{figure}

\paragraph{Model} This component serves as a unified interface for various MLLMs, handling model-specific API calls. It allows users to integrate from open-source models to proprietary APIs without altering the agent logic or evaluation protocol.

\paragraph{Task} This component encapsulates the specific navigation challenge and manage dataset splits. By treating the task as a modular input, our framework supports diverse navigation tasks within a unified evaluation protocol.

\paragraph{Agent} The Agent acts as the decision-making module that mediates the interaction between the MLLM and the Task. Its role is to embed environmental observations into structured prompts and parse the model's textual output into executable actions. To rigorously test different cognitive capabilities, we implement modular agent designs varying in memory and reasoning. We compare agents using text summarized action histories~\citep{zhou2024navgpt} against those building topological text maps~\citep{chen2024mapgpt} to test long-term spatial grounding. Similarly, to probe planning depth, we evaluate baselines ranging from direct prediction to those employing Chain-of-Thought (CoT)~\citep{wei2022chain} and self-reflection~\citep{yao2022react}. Detailed agent workflow and prompt templates are provided in the Appendix~\ref{app:prompt_design}.

% \begin{figure*}[t]
%     \centering
%     \includegraphics[width=\textwidth]{Fig/main.png}
%     \caption{Overview of the VLN-MME benchmark. \textbf{(Left)} The composition of the benchmark, detailing the diverse set of \textbf{Tasks}, \textbf{Agents}, and \textbf{Models} it supports. \textbf{(Right)} A statistical comparison of our benchmark's R2R data subset against the original R2R val\_unseen split}
%     \label{fig:vlnmme}
%     % , showing similar distributions for key metrics like instruction word count and path length.
% \end{figure*}

% The orchestration of these components is managed by a central \textbf{Runner} module, which uses an efficient configuration system for easy and reproducible experiment setup. The Runner handles the entire evaluation lifecycle. It begins by loading the pre-stored simulator-free environment, whose construction is detailed in Section~\ref{sec:simulator_free}. Concurrently, it dynamically loads the specified dataset and splits via the factory, as described in Section~\ref{sec:dataset-construction}. During an episode, the Runner acts as the low-level intermediary between the agent and the environment; it services agent requests for state information, renders observations, and executes actions. Throughout this process, the Runner logs all interactions for detailed post-hoc analysis. Upon completion of all episodes, it is responsible for calculating and reporting the final evaluation metrics. This centralized design cleanly separates high-level agent logic from low-level environment management, reinforcing the framework's modularity.

\paragraph{Extensibility and Orchestration} To ensure modularity, we adopt a unified \textbf{factory pattern} for instantiating all three component types. By registering each component with a unique identifier resolved at runtime, this design enables true ``plug-and-play'' extensibility: integrating a new agent or model requires only a simple registration, preserving the integrity of the core evaluation logic.

The evaluation lifecycle is coordinated by a central Runner module, which ensures reproducibility through a lightweight configuration system. The Runner manages the initialization of the simulator-free environment (Section~\ref{sec:simulator_free}) and the dynamic loading of datasets (Section~\ref{sec:dataset-construction}). During execution, it acts as the intermediary, facilitating the exchange of observations and actions between the agent and environment while logging interactions for granular post-hoc analysis. This architecture strictly separates high-level reasoning capabilities from low-level execution details, reinforcing the framework's modularity.

\subsection{Dataset Construction for Efficient Evaluation}
\label{sec:dataset-construction}

To address the computational challenges of evaluating large models on existing, large-scale VLN datasets and to facilitate rapid experimentation, we constructed a curated benchmark for efficient yet representative evaluation. Following the broader definition of VLN~\citep{zheng2023towards, zhou2025same}, our benchmark is composed of samples carefully drawn from the validation unseen splits of three main datasets: R2R~\citep{anderson2018r2r}, REVERIE~\citep{qi2020reverie}, and ObjectNav~\citep{batra2020objectnav}. Detailed descriptions of these datasets and the rationale behind their selection are provided in the Appendix~\ref{app:dataset_and_context}.

% \begin{figure}[htbp]
% 	\centering
%     \includegraphics[width=\columnwidth]{Fig/Performance Comparison.png}
% 	\caption{Comparison of model performance on full val\_unseen splits vs. our curated benchmark for R2R and REVERIE.}
% 	\label{fig:representativeness}
% \end{figure}
% \input{Tab/representativeness}

Our construction employs stratified sampling to preserve diversity across three key axes: \textbf{scene complexity}, \textbf{path difficulty}, and \textbf{linguistic richness}. For instance, in R2R and REVERIE, we stratify episodes by Matterport3D scan ID to capture environmental variance, then sample proportionally from path length bins to maintain the difficulty distribution. Linguistic diversity is ensured by randomly selecting one of the three available natural language instructions for each trajectory. For ObjectNav, we additionally enforce a balanced distribution of target object categories. This process yields a compact benchmark that statistically mirrors the characteristics of the original datasets. To further validate the fidelity of our constructed benchmark, we evaluated several previous methods from VLN specialist to Finetuned MLLM, on both the full val\_unseen splits and our curated benchmark for R2R and REVERIE (all the details are provided in Appendix~\ref{app:visual_construction}). The results reveal a strong correlation in performance. Key metrics such as Success Rate (SR) and Success weighted by Path Length (SPL) on our benchmark closely track the performance on the full splits, with deviations typically within a 2-3 percentage point margin. This close alignment confirms that our stratified sampling approach successfully captures the intrinsic difficulty and diversity of the original datasets, establishing our benchmark as a reliable and efficient proxy for full-scale MLLM evaluation.

% Our construction strategy employs a task-specific stratified sampling process designed to maintain diversity across three key axes: \textbf{scene complexity}, \textbf{path difficulty}, and \textbf{linguistic richness}. For instance, when constructing the R2R portion of our benchmark from the original 783 unique trajectories, the process begins by stratifying episodes based on their Matterport3D scan ID to ensure the selection reflects the original distribution of environments. Within each scan-based group, trajectories are then binned by their path length - a proxy for navigational difficulty - and sampled proportionally from each bin. Finally, to ensure linguistic variety, one of the three available natural language instructions is selected at random for each chosen trajectory. A similar stratified methodology, adapted to the unique characteristics of each task, was applied to create the benchmark data for REVERIE. For ObjectNav, we also consider object balance, ensuring that the sampled object navigation episodes maintain a balanced distribution of objects from the previous benchmark. This meticulous process ensures that our resulting benchmark, while significantly smaller, retains a comparable distribution of these core characteristics to the original datasets, as illustrated in Figure~\ref{fig:vlnmme}.

\subsection{Simulator-Free Environment Design}
\label{sec:simulator_free}

While high-fidelity simulators are essential for rendering, their computational overhead creates a bottleneck for real-time evaluation. To address this, we introduce a \textbf{simulator-free} mode that implements a strategic ``space-for-time'' trade-off, decoupling evaluation from the rendering engine by substituting expensive GPU operations with efficient disk I/O. As detailed in Table~\ref{tab:efficiency_comparison}, this approach drastically lowers the hardware barrier: by loading pre-cached images instead of full 3D scene geometry and textures, we reduce VRAM usage from $\sim$10GB to $\sim$1.7GB ($6\times$ lower), enabling research on consumer-grade hardware. Furthermore, replacing on-the-fly rendering with direct image retrieval accelerates observation access by nearly $9\times$ (0.14s to 0.016s), shortening total episode duration by 20--30 seconds. This efficiency is reinforced by our discrete navigation graph, which removes the need for continuous waypoint prediction typical of VLN-CE, thereby streamlining the evaluation loop without sacrificing 3D spatial assessment.

\begin{table}[h!]
\centering
\small
\caption{\textbf{Efficiency Comparison.} Resource usage and runtime of our simulator-free, pre-rendered approach versus Habitat, showing substantially lower memory consumption and faster evaluation.}
\label{tab:efficiency_comparison}
\resizebox{\columnwidth}{!}{%
\begin{tabular}{lccc}
\toprule
\textbf{Metric} & \textbf{Ours} & \textbf{Habitat} & \textbf{Delta} \\
\midrule
VRAM Usage & $\sim$1.7 GB & $\sim$10 GB & \textbf{5.9$\times$ Lower} \\
Obs. Access & $\sim$0.016s & $\sim$0.14s & \textbf{8.8$\times$ Faster} \\
Time / Step & $t$ & $t + 1.5$s & \textbf{1.5s Faster} \\
Time / Episode & $T$ & $T + 25$s & \textbf{$\sim$25s Faster} \\
\bottomrule
\end{tabular}%
}
\end{table}

The core of this design is a standardized visual representation optimized for MLLM perception. Rather than using distorted equirectangular projections, we capture the panoramic context as a set of four non-overlapping perspective images, each with a 90$^{\circ}$ Field of View. This format preserves visual fidelity and aligns better with the pre-training distributions of standard vision encoders. The rationale for this specific projection strategy is detailed further in Appendix~\ref{app:visual_construction}. Within these views, navigable directions are annotated with distinct numerical markers derived from the underlying connectivity graph, sorting candidates by their global heading to provide a consistent spatial reference.

To bridge the gap between visual features and symbolic reasoning, we augment the environment with semantic metadata. We employ GPT-5 to generate descriptive captions for both the general scene and specific navigation candidates. These descriptions serve as auxiliary inputs for agents that rely on semantic memory; the generation prompts and validation of these captions are documented in the Appendix~\ref{app:semantic_generation}. Finally, to ensure immediate accessibility, we host all pre-rendered assets on open-source platforms. Our framework manages the automatic retrieval of these artifacts, enabling users to evaluate without complex simulator installation.

%% file: Sec/experiment.tex
\section{Experiments}
\label{sec:experiments}

\subsection{Settings}
\paragraph{Evaluation Metrics.} In this work, we focus exclusively on the navigation component of all tasks, without considering object grounding in REVERIE. We adopt a standard set of navigation metrics to evaluate agent performance: (1) \textit{Trajectory Length} (TL), which measures the average path length in meters; (2) \textit{Navigation Error} (NE), the average distance between the agent’s final position and the goal location; (3) \textit{Success Rate} (SR), the percentage of episodes where the final location is within 3 meters of the target; (4) \textit{Oracle Success Rate} (OSR), the success rate assuming an optimal stopping policy; (5) \textit{Success weighted by Path Length} (SPL)~\citep{jain2019stay}, which combines success with path efficiency; (6) \textit{Normalized Dynamic Time Warping} (nDTW)~\citep{ilharco2019ndtw}, which measures the trajectory similarity to the ground truth path; and (7) \textit{Success weighted by normalized DTW} (SDTW), a combined metric capturing both goal-reaching and trajectory fidelity.

\input{Tab/main_table}

\paragraph{Implementation Details}
We evaluate two proprietary and four open-source MLLMs in a zero-shot setting: GPT5, Gemini2.5 Pro, Qwen2.5-VL-7B~\citep{bai2025qwen2}, InternVL3-2/8B~\citep{zhu2025internvl3}, LLaVA-One-Vision-7B~\citep{li2024llava}. These models are integrated into eight distinct agent configurations, categorized into two primary classes: agents using text summarization and text map as memory. Each class includes four variants: a baseline, one with CoT prompting, one with reflection-based reasoning, and one featuring both CoT and reflection. All open-source MLLMs are served using the vLLM backend~\citep{kwon2023efficient} to ensure efficient inference and memory management. We assess their performance on all the tasks in our benchmark. Additionally, we compare these zero-shot agents against previously finetuned VLN specialist and finetuned MLLM agents on the R2R and REVERIE tasks, evaluating performance across both the full dataset from prior evaluation methods and our benchmark. All experiments are conducted on a single NVIDIA A100 GPU with 40GB VRAM.

\subsection{Performance Analysis}
\label{app:performance}
We evaluate our zero-shot MLLM-based agents against prior state-of-the-art methods, analyzing performance across different model architectures, reasoning strategies, and navigation tasks.
\paragraph{Model Capabilities.}
Our results, detailed in Table~\ref{tab:benchmark_results}, indicate that proprietary models like GPT-5 and Gemini-2.5 Pro currently establish the upper bound for zero-shot navigation. However, among open-weights models, Qwen2.5-VL-7B demonstrates remarkable robustness, consistently outperforming peers such as LLaVA-OneVision and InternVL3-2B. For instance, in the baseline NavGPT configuration on the Fine-Grained task, Qwen2.5-VL-7B achieves a Success Rate (SR) of 27.5\%, substantially surpassing LLaVA-OneVision (11.5\%) and InternVL3-2B (13.5\%). This positions Qwen2.5-VL as a strong and capable baseline for the open-source embodied community.
\paragraph{Agent Architecture and Reasoning Strategies.}
Counterintuitively, the integration of advanced prompting strategies like CoT or reflection does not consistently yield performance improvements and often proves detrimental. As shown in Table~\ref{tab:benchmark_results}, applying CoT to Qwen2.5-VL-7B in the Fine-Grained task drops the SR from 27.5\% to 21\%. Similarly, the distinction between \textit{Text Summarization} and \textit{Text Map} memory architectures is less critical than expected. Performance variances are generally minimal, although specific smaller models, such as InternVL3-2B, derive a tangible benefit from structured map memory in Coarse-grained tasks, improving from 7.33\% SR to 12\% SR.
We argue that these counterintuitive limitations stem from the models' deficient \textit{embodied context awareness}. Our investigation reveals two critical, interrelated issues. First, MLLMs exhibit a strong bias toward ``local reasoning'', relying heavily on immediate visual inputs and current instructions while neglecting the rich historical context. Second, due to this lack of historical grounding, the agents struggle to anticipate the downstream consequences of their actions, failing to adapt or recover from errors within the long-term sequential flow of the task. The comprehensive error analysis in Section~\ref{sec:discussion} provides further evidence supporting this conclusion.
% \paragraph{Zero-Shot vs. Finetuned Baselines.}
% A significant performance gap remains between our zero-shot agents and fully finetuned baselines. As referenced in Table~\ref{tab:compare_vlm}, state-of-the-art finetuned agents achieve significantly higher success rates (e.g., 
% 72\% SR on R2R) compared to the zero-shot average. However, zero-shot agents demonstrate non-trivial navigation capabilities, establishing a crucial and promising baseline for generalization without task-specific training. Furthermore, we observe that the performance of prior methods on our benchmark subset aligns consistently with their results on the full validation set, verifying the representativeness of our evaluation.
\paragraph{Task Difficulty Hierarchy.}
Finally, the results reveal a clear difficulty hierarchy across navigation categories. Object-Oriented Navigation proves the most tractable, with agents consistently achieving higher success rates. Fine-Grained Navigation presents moderate difficulty, while Coarse-Grained Navigation emerges as the most challenging task for MLLMs. This suggests that interpreting high-level instructions to navigate towards less precisely defined locations remains a substantial hurdle for current zero-shot reasoning.

\begin{figure}[h]
    \centering
    \includegraphics[width=\columnwidth]{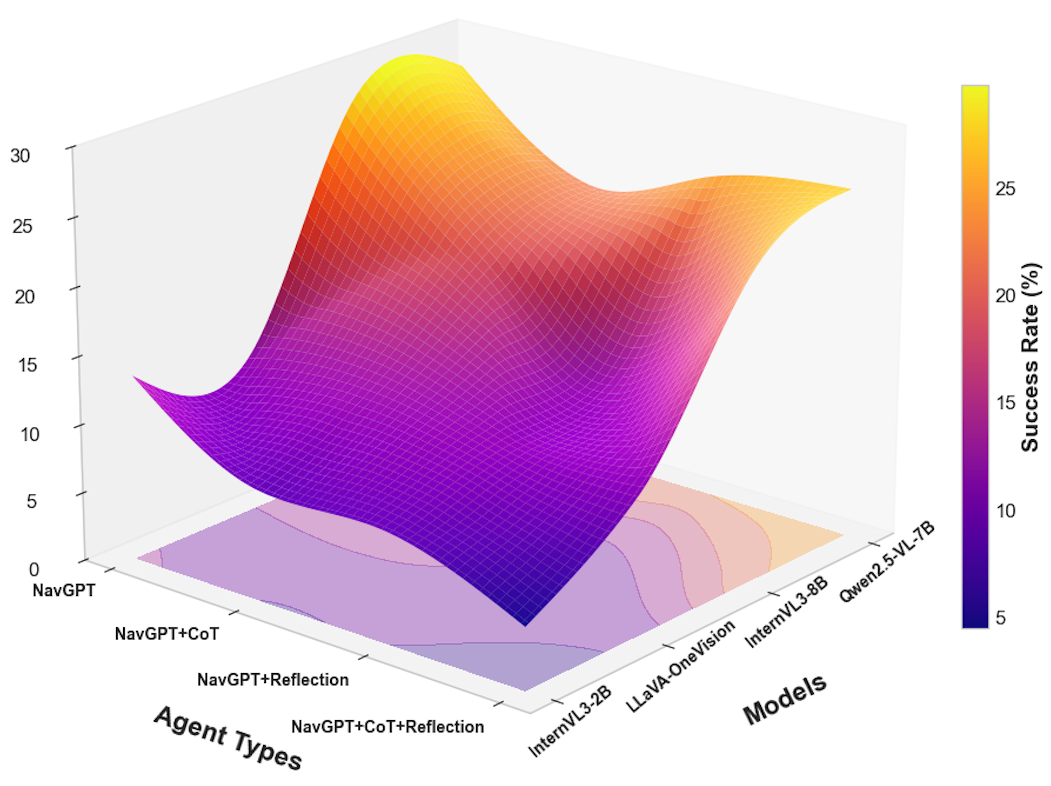}
    \caption{Performance comparison of agents using text summarization memory under different reasoning strategies across multiple backbone MLLMs.}
    \label{fig:agent and reasoning compare}
\end{figure}

\subsection{Discussion}

\label{sec:discussion}

\begin{figure}[h]
    \centering
    \includegraphics[width=\columnwidth]{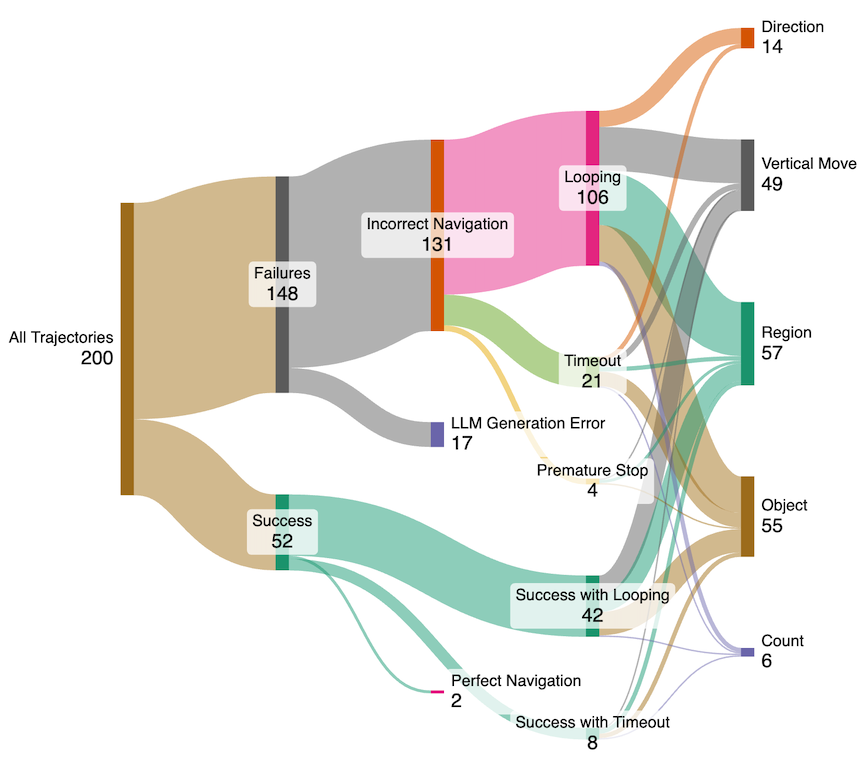}
    \caption{A high-level analysis of success and failure modes for Qwen2.5-VL-7B model using an agent with text map memory.}
    \label{fig:error}
\end{figure}

As discussed in section~\ref{app:performance}, we reveal some counterfactual behavior when MLLMs perform embodied navigation. We further conduct an error analysis to understand their error pattern and find that they are hindered by fundamental limitations across several cognitive dimensions (see more details in Appendix~\ref{app:error_analysis}). Interestingly, we find that the high navigation failure rate is overwhelmingly dominated by looping behaviors, shown in Figure~\ref{fig:error}. It is not a superficial issue but symptomatic of deeper challenges in instruction fidelity, spatial reasoning, historical context utilization, and the grounding of multi-modal perception into action. We discuss these three interconnected aspects below.

% To demonstrate the practical value of these observations, we leveraged our error analysis to conduct a diagnostic study on hard negatives (detailed in Appendix {\color{red}F}). The results confirm that addressing these identified reasoning deficits leads to significant performance recovery, validating the utility of our analysis. 

% As discussed in Section~\ref{app:performance}, MLLMs exhibit counterfactual navigation behaviors. Our error analysis (Appendix~{\color{red}E}) further reveals that failures are dominated by looping (Figure~\ref{fig:error}), which reflects deeper deficits in instruction fidelity, spatial reasoning, and context grounding rather than superficial errors. A diagnostic study on hard negatives (Appendix~{\color{red}F}) confirms that addressing these reasoning gaps yields significant performance recovery, validating our analysis. We discuss these three interconnected aspects below.

\paragraph{Instruction Following and Reasoning Fidelity.}
A primary challenge is the limited fidelity with which MLLMs adhere to complex instructions, particularly those governing their reasoning process. While the models can follow basic output formatting prompts, they struggle with more abstract meta-instructions. For instance, when prompted with CoT or reflection mechanisms to explicitly ``reason based on history and the map," the agents often diverge, reverting to a reactive, myopic reasoning pattern that ignores the very context they were instructed to use. This disconnect helps explain why adding CoT and reflection did not consistently improve performance (Table~\ref{tab:benchmark_results}); the models did not faithfully execute the intended reasoning strategy. This suggests a significant gap between simply conditioning a model on a prompt and instilling a robust, procedural reasoning capability.

\paragraph{Spatial and Environmental Understanding.}
Our fine-grained error analysis reveals that profound weaknesses in spatial understanding are the root cause of most navigational failures. Of 131 errors analyzed, a staggering 106 were due to persistent looping, a direct consequence of the model's inability to ground instructions in the 3D environment. This manifests in specific, recurring issues like poor region recognition, failure to reason about verticality on stairs, and basic directional confusion. The fact that providing an explicit topological map failed to yield significant gains highlights a deeper problem: the agent cannot connect abstract spatial knowledge to its visual perception and actions. Furthermore, the agent critically fails at sequential decision-making, which is essential for navigation. The rampant looping behavior clearly shows that the agent does not learn from its trajectory to avoid repeating mistakes. This is not a problem of memory capacity, as the history rarely exceeds the model's context window, but rather one of memory utilization. The model has access to its past actions but cannot ground its current decisions in that history to self-correct. In fact, the observation that simpler history formats can outperform complex ones suggests that too much historical information creates a cognitive load, confusing the agent instead of guiding it.

\paragraph{Perception-Action Grounding.}
Finally, we observe a critical gap between multimodal perception and embodied action. The MLLM's visual grounding is functional at a recognition level; for example, it can often correctly identify a ``staircase" or a target ``chair" in its textual reasoning trace. This indicates that the visual and language modalities are connected. However, this recognition consistently fails to translate into correct action. The agent sees the stairs but walks past them in a loop. It may even get very close to the goal, demonstrating it has successfully grounded the target object visually, yet fails to execute the final `STOP' action. This is powerfully illustrated by our success-case analysis, where most successful episodes involved inefficient looping near the target before stopping. This ``perception-action gap" shows that the greatest challenge for MLLMs in VLN is not just seeing and describing the world, but effectively acting within it.

\subsection{Diagnostic Case Study}
\label{app:diagnostic_study}
To demonstrate the utility of our benchmark as a scientific tool for uncovering MLLM capabilities, we conducted an in-depth investigation on a specific subset of ``Hard Negatives." This subset consists of 25 trajectories from the fine-grained task where all open-source models consistently failed using the standard text-summary memory agent. We revisited these failure cases using Qwen2.5VL-7B as the primary navigator and designed two experimental settings to isolate the source of the failure:

\paragraph{(1) Oracle-Guided Navigation.} 
To test if the failure was due to a lack of visual understanding or a lack of high-level reasoning, we introduced a stronger model (Qwen3VL-4B~\citep{yang2025qwen3}) acting as an Oracle Assistant. When the navigator struggled (\textit{e.g.}, looping, moving in a critically wrong direction, or entering the wrong region), it was allowed to query the Oracle. The Oracle provided high-level reasoning guidance and suggested potential plans and  actions. The navigator achieved a significant success rate with this reasoning support as shown in Table~\ref{tab:oracle_guidance}, suggesting that the base model possesses the fundamental navigational capability but lacks the high-level planning or error-correction logic required for these hard cases.

\begin{table}[h!]
    \centering
    \caption{Effect of Oracle Assistance on Hard Negatives.}
    \label{tab:oracle_guidance}
    \vspace{1mm}
    \resizebox{\columnwidth}{!}{%
    \begin{tabular}{l c c c}
        \toprule
        \textbf{Method} & \textbf{SR} $\uparrow$ & \textbf{OSR} $\uparrow$ & \textbf{SPL} $\uparrow$ \\
        \midrule
        Baseline (Qwen2.5VL-7B) & 0.00 & 0.00 & 0.00 \\
        + Oracle Assistant (Qwen3VL) & \textbf{52.00} & \textbf{68.00} & \textbf{41.28} \\
        \bottomrule
    \end{tabular}%
    }
\end{table}

\paragraph{(2) Failure-Aware In-Context Learning.} 
To test if the model could self-correct given awareness of common pitfalls, we replaced the zero-shot evaluation with a few-shot setup. We provided the model with $N$ examples of potential failure cases identified in our error analysis ($N=\{1, 2, 3\}$). As shown in Table~\ref{tab:failure_icl}, providing these ``negative'' examples in the prompt yielded a performance increase. However, the gains were modest compared to the Oracle intervention, indicating that while awareness of failure modes helps, active reasoning support is more critical for solving complex navigation challenges.

\begin{table}[h!]
    \centering
    \caption{Effect of Failure-Aware Few-Shot Prompting on Hard Negatives.}
    \label{tab:failure_icl}
    \vspace{1mm}
    \resizebox{\columnwidth}{!}{%
    \begin{tabular}{l c c c}
        \toprule
        \textbf{Method} & \textbf{SR} $\uparrow$ & \textbf{OSR} $\uparrow$ & \textbf{SPL} $\uparrow$ \\
        \midrule
        Zero-shot (Baseline) & 0.00 & 0.00 & 0.00 \\
        1-shot Failure Example & 12.00 & 24.00 & 9.42 \\
        2-shot Failure Examples & 16.00 & 28.00 & 11.25 \\
        3-shot Failure Examples & 16.00 & \textbf{32.00} & \textbf{14.29} \\
        \bottomrule
    \end{tabular}%
    }
\end{table}

%% file: Tab/main_table.tex
\definecolor{tableheadgray}{gray}{0.9}
\definecolor{Gray}{gray}{0.94}
\begin{table*}[!t]
\centering
\resizebox{\textwidth}{!}{
\setlength{\tabcolsep}{2pt}
\begin{tabular}{@{}l|cccccc>{\columncolor{Gray}}c>{\columncolor{Gray}}c|cccccc>{\columncolor{Gray}}c>{\columncolor{Gray}}c|cccccc>{\columncolor{Gray}}c>{\columncolor{Gray}}c@{}}
\toprule
\multicolumn{1}{c|}{\multirow{2}{*}{\textbf{Agent / MLLM}}} & \multicolumn{8}{c|}{\textbf{Fine-Grained Navigation}} & \multicolumn{8}{c|}{\textbf{Coarse-grained Navigation}} & \multicolumn{8}{c}{\textbf{Object-Oriented Navigation}} \\
\cmidrule(lr){2-9} \cmidrule(lr){10-17} \cmidrule(lr){18-25}
\multicolumn{1}{c|}{} & \textbf{TL} & \textbf{NE} & \textbf{nDTW} & \textbf{SDTW} & \textbf{CLS} & \textbf{OSR} & \multicolumn{1}{c}{\textbf{SR}} & \multicolumn{1}{c|}{\textbf{SPL}} & \textbf{TL} & \textbf{NE} & \textbf{nDTW} & \textbf{SDTW} & \textbf{CLS} & \textbf{OSR} & \multicolumn{1}{c}{\textbf{SR}} & \multicolumn{1}{c|}{\textbf{SPL}} & \textbf{TL} & \textbf{NE} & \textbf{nDTW} & \textbf{SDTW} & \textbf{CLS} & \textbf{OSR} & \multicolumn{1}{c}{\textbf{SR}} & \multicolumn{1}{c}{\textbf{SPL}} \\
\midrule
\rowcolor{Cerulean!20} \multicolumn{25}{c}{\textbf{Text Summarization Memory Agents}} \\
\midrule
\rowcolor{tableheadgray} \multicolumn{25}{l}{\textbf{NavGPT}} \\
\midrule
GPT-5 & 9.12 & \textbf{6.28} & \textbf{37.96} & 22.02 & \textbf{34.11} & 57.50 & 38.50 & 29.23 & 9.37 & \textbf{8.03} & \textbf{27.11} & 14.88 & 21.05 & 49.50 & 30.00 & 20.76 & 9.63 & \textbf{4.41} & \textbf{26.52} & 16.33 & 20.13 & \textbf{79.00} & 48.00 & 23.84 \\
Gemini-2.5 Pro & \textbf{8.94} & 6.17 & 38.07 & \textbf{22.43} & 34.62 & \textbf{60.50} & \textbf{41.00} & \textbf{32.67} & \textbf{9.21} & 7.89 & 27.47 & \textbf{15.13} & \textbf{21.38} & \textbf{52.00} & \textbf{33.50} & \textbf{24.38} & \textbf{9.48} & 4.34 & 26.78 & \textbf{16.54} & \textbf{20.43} & \textbf{79.00} & \textbf{51.50} & \textbf{27.19} \\
InternVL3-2B & 9.89 & 8.56 & 21.25 & 5.75 & 21.59 & 27.00 & 13.50 & 5.46 & 10.13 & 10.18 & 15.30 & 2.97 & 17.17 & 16.33 & 7.33 & 2.50 & 10.35 & 6.27 & 13.67 & 2.55 & 15.57 & 40.50 & 21.50 & 3.57 \\
InternVL3-8B & 11.74 & 7.55 & 25.28 & 13.38 & 26.22 & 50.50 & 28.00 & 12.61 & 11.90 & 9.25 & 17.32 & 8.18 & 18.84 & 30.67 & 20.00 & 7.18 & 11.55 & 4.63 & 14.51 & 5.09 & 17.36 & 56.00 & 39.00 & 7.69 \\
LLaVA-OV-7B & 8.04 & 8.40 & 26.34 & 5.70 & 25.53 & 20.50 & 11.50 & 4.94 & 9.85 & 9.35 & 19.27 & 6.09 & 18.39 & 20.00 & 14.67 & 5.19 & 9.52 & 5.93 & 16.54 & 5.11 & 14.71 & 41.00 & 27.50 & 4.51 \\
Qwen2.5-VL-7B & 8.54 & 6.99 & 35.97 & 18.88 & 34.85 & 44.00 & 27.50 & 17.11 & 8.94 & 8.55 & 23.79 & 9.88 & 23.55 & 27.33 & 18.67 & 9.00 & 9.07 & 4.65 & 21.83 & 10.63 & 23.52 & 56.50 & 37.50 & 13.18 \\
\midrule
\rowcolor{tableheadgray} \multicolumn{25}{l}{\textbf{NavGPT w/ CoT}} \\
\midrule
GPT-5 & 9.47 & 6.42 & \textbf{39.17} & \textbf{18.89} & \textbf{32.54} & \textbf{54.50} & \textbf{36.00} & \textbf{27.31} & 9.71 & 8.24 & \textbf{32.53} & \textbf{12.04} & \textbf{26.82} & \textbf{46.00} & \textbf{28.50} & \textbf{20.42} & 9.92 & 4.58 & \textbf{31.06} & \textbf{14.03} & \textbf{25.07} & \textbf{70.00} & \textbf{46.50} & \textbf{23.67} \\
Gemini-2.5 Pro & \textbf{9.31} & \textbf{6.29} & 39.64 & 19.12 & 32.86 & 50.00 & 32.50 & 23.88 & \textbf{9.56} & \textbf{8.11} & 32.83 & 12.22 & 27.14 & 41.50 & 24.00 & 16.93 & \textbf{9.76} & \textbf{4.49} & 31.38 & 14.24 & 25.31 & 65.50 & 42.00 & 19.27 \\
InternVL3-2B & 6.21 & 8.84 & 26.15 & 4.28 & 27.45 & 14.50 & 8.00 & 4.47 & 4.98 & 9.87 & 23.39 & 3.11 & 25.09 & 9.00 & 5.33 & 3.24 & 6.78 & 5.56 & 25.51 & 5.34 & 23.98 & 41.00 & 25.00 & 6.66 \\
InternVL3-8B & 9.07 & 7.56 & 29.22 & 10.90 & 29.59 & 35.50 & 19.00 & 10.95 & 7.96 & 9.39 & 24.93 & 9.14 & 26.40 & 22.00 & 15.33 & 9.31 & 6.43 & 5.31 & 27.22 & 10.30 & 27.35 & 43.50 & 34.50 & 12.67 \\
LLaVA-OV-7B & 7.97 & 8.85 & 23.20 & 5.60 & 24.43 & 22.50 & 12.50 & 5.41 & 8.43 & 9.47 & 21.09 & 7.13 & 21.38 & 20.33 & 14.00 & 5.90 & 9.69 & 5.66 & 15.90 & 6.24 & 17.92 & 47.50 & 33.50 & 7.31 \\
Qwen2.5-VL-7B & 9.04 & 7.97 & 30.23 & 11.56 & 31.36 & 37.50 & 21.00 & 11.41 & 8.29 & 9.85 & 24.02 & 8.82 & 25.59 & 24.33 & 15.67 & 8.10 & 6.86 & 5.13 & 28.63 & 12.54 & 28.32 & 44.50 & 33.00 & 13.25 \\
\midrule
\rowcolor{tableheadgray} \multicolumn{25}{l}{\textbf{NavGPT w/ Reflection}} \\
\midrule
GPT-5 & 9.61 & 6.37 & \textbf{40.04} & 19.53 & \textbf{33.14} & 50.50 & 33.00 & 24.18 & 9.82 & 8.13 & \textbf{33.03} & 12.51 & \textbf{27.08} & 43.00 & 25.00 & 17.34 & 10.03 & 4.47 & \textbf{30.02} & 15.06 & 24.09 & 68.00 & 43.00 & 20.62 \\
Gemini-2.5 Pro & \textbf{9.44} & \textbf{6.23} & 40.39 & \textbf{19.71} & 33.47 & \textbf{55.00} & \textbf{37.50} & \textbf{28.73} & \textbf{9.67} & \textbf{7.98} & 33.34 & \textbf{12.72} & 27.33 & \textbf{47.50} & \textbf{29.00} & \textbf{21.68} & \textbf{9.88} & \textbf{4.38} & 30.27 & \textbf{15.21} & \textbf{24.34} & \textbf{72.50} & \textbf{47.50} & \textbf{25.14} \\
InternVL3-2B & 6.50 & 8.94 & 29.91 & 5.00 & 30.13 & 16.00 & 8.00 & 5.20 & 7.20 & 9.41 & 25.49 & 4.60 & 24.93 & 15.00 & 8.50 & 4.80 & 6.43 & 5.56 & 26.53 & 6.14 & 25.28 & 43.00 & 28.00 & 7.83 \\
InternVL3-8B & 4.54 & 7.99 & 34.25 & 8.89 & 34.44 & 19.00 & 12.00 & 9.53 & 6.82 & 10.20 & 23.58 & 6.04 & 27.79 & 17.33 & 11.00 & 5.97 & 8.35 & 5.02 & 18.79 & 6.68 & 22.14 & 51.00 & 32.50 & 9.12 \\
LLaVA-OV-7B & 2.81 & 8.01 & 38.17 & 8.43 & 38.39 & 11.00 & 10.50 & 9.44 & 5.15 & 9.34 & 28.04 & 5.91 & 30.22 & 14.67 & 9.33 & 5.82 & 9.35 & 5.58 & 15.48 & 6.11 & 17.05 & 47.50 & 34.00 & 7.90 \\
Qwen2.5-VL-7B & 6.93 & 7.17 & 36.51 & 16.44 & 33.59 & 32.50 & 24.00 & 14.95 & 6.96 & 8.76 & 26.88 & 7.97 & 25.78 & 18.00 & 12.00 & 7.97 & 7.55 & 5.06 & 23.33 & 11.04 & 25.23 & 50.50 & 35.50 & 14.67 \\
\midrule
\rowcolor{tableheadgray} \multicolumn{25}{l}{\textbf{NavGPT w/ CoT \& Reflection}} \\
\midrule
GPT-5 & 9.52 & 6.31 & \textbf{39.56} & \textbf{20.08} & \textbf{33.02} & \textbf{56.50} & \textbf{38.50} & \textbf{29.81} & 9.77 & 8.18 & \textbf{32.04} & \textbf{13.07} & \textbf{26.03} & \textbf{48.50} & \textbf{30.00} & \textbf{22.17} & 9.97 & \textbf{4.36} & \textbf{29.08} & \textbf{15.51} & \textbf{23.06} & \textbf{73.50} & \textbf{48.50} & \textbf{25.92} \\
Gemini-2.5 Pro & \textbf{9.36} & \textbf{6.19} & 39.83 & 20.24 & 33.27 & 52.00 & 34.00 & 25.74 & \textbf{9.61} & \textbf{8.04} & 32.32 & 13.21 & 26.24 & 43.50 & 25.50 & 18.36 & \textbf{9.81} & 4.27 & 29.33 & 15.66 & 23.22 & 68.00 & 43.50 & 21.28 \\
InternVL3-2B & 7.15 & 9.24 & 22.45 & 2.05 & 23.47 & 15.00 & 4.50 & 1.70 & 7.30 & 9.78 & 22.33 & 5.30 & 24.07 & 15.00 & 9.33 & 4.63 & 6.94 & 6.25 & 21.29 & 6.11 & 21.47 & 37.50 & 24.50 & 7.26 \\
InternVL3-8B & 7.22 & 7.47 & 36.62 & 16.18 & 35.98 & 32.50 & 22.00 & 15.33 & 8.95 & 9.07 & 24.11 & 10.24 & 25.78 & 28.67 & 17.33 & 10.07 & 9.18 & 5.30 & 18.59 & 5.99 & 21.45 & 51.50 & 32.50 & 8.14 \\
LLaVA-OV-7B & 7.61 & 8.48 & 28.01 & 6.31 & 26.32 & 22.00 & 10.00 & 5.83 & 8.44 & 8.68 & 24.73 & 8.37 & 22.60 & 22.00 & 14.00 & 6.78 & 8.55 & 5.66 & 21.26 & 6.74 & 19.69 & 44.00 & 28.50 & 7.25 \\
Qwen2.5-VL-7B & 7.82 & 7.53 & 34.86 & 17.65 & 34.80 & 38.50 & 25.50 & 17.68 & 7.19 & 9.48 & 27.07 & 7.81 & 28.22 & 18.00 & 11.67 & 7.89 & 7.60 & 5.39 & 26.52 & 11.49 & 26.98 & 47.00 & 36.00 & 13.67 \\
\midrule
\rowcolor{Cerulean!20} \multicolumn{25}{c}{\textbf{Text Map Memory Agents}} \\
\midrule
\rowcolor{tableheadgray} \multicolumn{25}{l}{\textbf{MapGPT}} \\
\midrule
GPT-5 & 9.32 & 6.21 & \textbf{39.03} & 21.08 & \textbf{33.05} & 52.50 & 34.00 & 25.83 & 9.53 & 7.94 & \textbf{29.07} & 13.02 & 23.04 & 45.00 & 26.00 & 18.29 & 9.73 & 4.33 & \textbf{28.06} & 16.08 & 22.01 & 74.00 & 44.00 & 20.91 \\
Gemini-2.5 Pro & \textbf{9.17} & \textbf{6.09} & 39.34 & \textbf{21.22} & 33.23 & \textbf{57.50} & \textbf{39.50} & \textbf{30.72} & \textbf{9.38} & \textbf{7.81} & 29.26 & \textbf{13.17} & \textbf{23.21} & \textbf{50.00} & \textbf{31.50} & \textbf{23.16} & \textbf{9.58} & \textbf{4.26} & 28.24 & \textbf{16.19} & \textbf{22.13} & \textbf{79.50} & \textbf{49.50} & \textbf{26.24} \\
InternVL3-2B & 9.84 & 8.61 & 20.18 & 3.85 & 21.89 & 22.50 & 11.00 & 3.71 & 10.19 & 9.59 & 18.10 & 5.24 & 19.97 & 19.00 & 12.00 & 4.41 & 10.35 & 5.93 & 13.59 & 3.69 & 15.57 & 46.50 & 27.50 & 4.41 \\
InternVL3-8B & 6.78 & 7.70 & 34.34 & 13.06 & 33.78 & 32.00 & 18.00 & 12.46 & 7.26 & 9.16 & 26.63 & 8.04 & 27.62 & 22.33 & 13.67 & 7.87 & 5.95 & 5.26 & 28.03 & 8.98 & 27.39 & 44.50 & 31.50 & 11.61 \\
LLaVA-OV-7B & 4.97 & 8.44 & 31.70 & 5.29 & 32.62 & 15.50 & 8.50 & 5.59 & 8.58 & 8.96 & 23.33 & 7.33 & 22.39 & 22.00 & 14.67 & 6.48 & 7.92 & 5.86 & 20.66 & 4.63 & 18.34 & 35.50 & 22.50 & 4.28 \\
Qwen2.5-VL-7B & 8.16 & 7.13 & 34.31 & 17.39 & 33.37 & 38.00 & 26.00 & 17.31 & 10.52 & 8.53 & 21.27 & 10.85 & 20.78 & 32.33 & 21.67 & 8.96 & 9.77 & 4.82 & 20.06 & 9.06 & 22.13 & 52.50 & 36.50 & 11.05 \\
\midrule
\rowcolor{tableheadgray} \multicolumn{25}{l}{\textbf{MapGPT w/ CoT}} \\
\midrule
GPT-5 & 9.62 & 6.51 & \textbf{40.09} & \textbf{17.04} & \textbf{32.08} & \textbf{50.00} & \textbf{32.50} & \textbf{27.14} & 9.83 & 8.31 & \textbf{34.05} & \textbf{10.06} & \textbf{28.03} & \textbf{42.50} & \textbf{24.50} & \textbf{19.68} & 9.93 & 4.67 & \textbf{32.08} & \textbf{13.01} & \textbf{26.02} & \textbf{68.50} & \textbf{44.50} & \textbf{23.21} \\
Gemini-2.5 Pro & \textbf{9.46} & \textbf{6.38} & 40.36 & 17.17 & 32.24 & 48.50 & 31.00 & 23.37 & \textbf{9.68} & \textbf{8.19} & 34.23 & 10.14 & 28.22 & 40.00 & 23.00 & 15.84 & \textbf{9.78} & \textbf{4.59} & 32.27 & 13.16 & 26.21 & 62.50 & 39.50 & 18.76 \\
InternVL3-2B & 5.09 & 8.99 & 26.43 & 3.94 & 27.62 & 14.00 & 6.50 & 4.12 & 5.22 & 9.88 & 23.01 & 1.77 & 24.48 & 8.00 & 4.00 & 1.58 & 6.66 & 6.31 & 21.14 & 4.69 & 21.05 & 35.50 & 20.00 & 4.76 \\
InternVL3-8B & 5.96 & 8.79 & 31.15 & 8.64 & 33.11 & 22.50 & 12.00 & 9.66 & 5.82 & 9.07 & 30.01 & 9.09 & 30.77 & 17.33 & 13.33 & 8.77 & 6.55 & 5.34 & 26.74 & 9.38 & 27.22 & 46.50 & 34.00 & 12.33 \\
LLaVA-OV-7B & 8.34 & 8.96 & 19.61 & 2.84 & 19.10 & 13.50 & 8.50 & 2.83 & 7.82 & 9.63 & 20.22 & 4.10 & 21.21 & 13.00 & 8.67 & 3.38 & 7.64 & 6.43 & 19.11 & 3.21 & 18.63 & 31.00 & 16.00 & 4.14 \\
Qwen2.5-VL-7B & 7.45 & 7.78 & 30.38 & 10.86 & 29.67 & 28.50 & 17.00 & 10.47 & 6.48 & 8.49 & 30.81 & 11.11 & 31.34 & 21.00 & 16.33 & 10.61 & 9.51 & 4.78 & 22.14 & 8.65 & 23.29 & 54.00 & 32.00 & 9.95 \\
\midrule
\rowcolor{tableheadgray} \multicolumn{25}{l}{\textbf{MapGPT w/ Reflection}} \\
\midrule
GPT-5 & 9.72 & 6.41 & \textbf{41.02} & \textbf{18.09} & \textbf{33.03} & \textbf{54.00} & \textbf{36.50} & \textbf{28.19} & 9.91 & 8.26 & \textbf{35.06} & \textbf{11.04} & \textbf{29.07} & \textbf{42.50} & \textbf{25.50} & \textbf{20.73} & 10.02 & 4.62 & \textbf{31.09} & \textbf{14.05} & \textbf{25.08} & \textbf{69.50} & \textbf{45.50} & \textbf{23.65} \\
Gemini-2.5 Pro & \textbf{9.57} & \textbf{6.28} & 41.28 & 18.16 & 33.21 & 49.50 & 32.00 & 24.31 & \textbf{9.77} & \textbf{8.14} & 35.27 & 11.13 & 29.25 & 41.50 & 24.00 & 16.88 & \textbf{9.87} & \textbf{4.54} & 31.24 & 14.17 & 25.23 & 64.00 & 41.00 & 19.94 \\
InternVL3-2B & 2.37 & 8.55 & 32.83 & 2.97 & 33.54 & 6.00 & 4.00 & 3.26 & 2.45 & 9.58 & 27.97 & 2.81 & 31.14 & 5.00 & 3.67 & 3.31 & 9.72 & 5.85 & 15.52 & 3.04 & 17.39 & 44.00 & 25.00 & 4.50 \\
InternVL3-8B & 5.85 & 7.80 & 36.05 & 11.48 & 35.18 & 28.50 & 16.50 & 10.89 & 6.49 & 8.80 & 27.19 & 6.91 & 28.24 & 19.33 & 12.67 & 6.64 & 6.23 & 5.50 & 26.92 & 8.12 & 25.74 & 40.50 & 30.00 & 10.36 \\
LLaVA-OV-7B & 8.35 & 8.50 & 26.62 & 5.50 & 23.33 & 20.00 & 10.00 & 5.50 & 8.20 & 9.47 & 25.84 & 6.00 & 25.16 & 19.00 & 11.00 & 6.00 & 9.46 & 6.11 & 15.90 & 2.53 & 15.57 & 33.50 & 15.50 & 2.02 \\
Qwen2.5-VL-7B & 10.41 & 7.12 & 27.97 & 12.88 & 25.38 & 41.00 & 26.50 & 10.12 & 9.60 & 8.67 & 23.29 & 7.62 & 20.63 & 23.67 & 15.67 & 6.00 & 10.15 & 4.90 & 17.18 & 6.33 & 17.34 & 46.00 & 33.50 & 7.41 \\
\midrule
\rowcolor{tableheadgray} \multicolumn{25}{l}{\textbf{MapGPT w/ CoT \& Reflection}} \\
\midrule
GPT-5 & 9.82 & 6.59 & \textbf{42.07} & \textbf{16.04} & \textbf{34.09} & \textbf{52.00} & \textbf{34.50} & \textbf{26.13} & 10.03 & 8.39 & \textbf{36.05} & \textbf{9.03} & \textbf{30.02} & \textbf{40.50} & \textbf{24.50} & \textbf{18.67} & 10.11 & 4.76 & \textbf{33.08} & \textbf{12.07} & \textbf{27.03} & \textbf{66.50} & \textbf{42.50} & \textbf{21.79} \\
Gemini-2.5 Pro & \textbf{9.66} & \textbf{6.46} & 42.26 & 16.11 & 34.24 & 47.50 & 30.00 & 22.28 & \textbf{9.88} & \textbf{8.27} & 36.28 & 9.16 & 30.17 & 39.50 & 22.00 & 15.41 & \textbf{9.96} & \textbf{4.68} & 33.27 & 12.12 & 27.16 & 61.00 & 38.00 & 17.86 \\
InternVL3-2B & 7.52 & 8.82 & 23.45 & 4.98 & 24.08 & 20.00 & 9.00 & 4.88 & 7.35 & 9.95 & 19.60 & 2.02 & 21.94 & 12.67 & 4.67 & 1.61 & 7.26 & 6.15 & 22.73 & 3.77 & 22.14 & 35.50 & 18.00 & 4.74 \\
InternVL3-8B & 8.54 & 8.42 & 28.16 & 10.84 & 29.42 & 34.50 & 18.00 & 10.84 & 9.64 & 9.76 & 20.36 & 6.75 & 22.16 & 24.00 & 13.00 & 6.06 & 8.95 & 5.38 & 16.44 & 5.85 & 19.04 & 53.50 & 33.50 & 8.01 \\
LLaVA-OV-7B & 8.35 & 8.50 & 25.55 & 6.04 & 25.47 & 23.00 & 13.00 & 5.81 & 8.14 & 9.56 & 21.74 & 4.86 & 21.98 & 17.67 & 9.00 & 4.53 & 8.53 & 5.74 & 21.01 & 4.35 & 20.49 & 37.00 & 24.50 & 5.45 \\
Qwen2.5-VL-7B & 6.29 & 9.07 & 25.82 & 7.03 & 25.41 & 20.50 & 14.00 & 7.12 & 7.09 & 8.94 & 25.66 & 7.27 & 25.81 & 18.67 & 10.67 & 6.68 & 7.45 & 5.90 & 22.67 & 5.23 & 21.56 & 36.50 & 25.50 & 6.78 \\
\bottomrule
\end{tabular}%
}
\caption{Performance Comparison of MLLM-based Agents on \ours. Agents are grouped by their primary architecture type. Best performance per group is marked in bold.}
\label{tab:benchmark_results}
\end{table*}

%% file: Sec/conclusion.tex
\section{Conclusion}
\label{sec:conclusion}

In this work, we introduce VLN-MME, a modular, simulator-free framework for diagnosing MLLMs as zero-shot embodied agents. Our evaluation reveals a critical "perception-action gap": models possess robust visual grounding yet fail at sequential decision-making, with Chain-of-Thought strategies often proving detrimental due to limited context awareness. Crucially, our diagnostic study on hard negatives confirms that these failures stem primarily from deficient strategic planning rather than perceptual limitations, as evidenced by substantial performance recovery under oracle guidance. Ultimately, VLN-MME highlights that future MLLM development must prioritize bridging the gap between static visual understanding and dynamic, history-aware strategic reasoning.

%% file: Sec/limitation.tex
\section*{Limitation}
We present and discuss the limitations of our work to outline the scope of our analysis. First, due to computational and temporal constraints, our evaluation covers a representative selection of MLLMs and agent designs rather than an exhaustive spectrum. However, we mitigate this by releasing a highly modular interface that enables the community to seamlessly integrate and benchmark emerging models. Second, while our framework supports broader capabilities such as dialog-based navigation and multi-language instructions, we restricted our current analysis to standard instruction following to establish a rigorous baseline. We designed the dataset construction pipeline to be extensible, allowing future researchers to easily incorporate these diverse tasks. Finally, this work is primarily diagnostic: we identify critical deficiencies in MLLM 3D spatial reasoning and strategic planning but do not propose specific algorithmic remedies. We posit these findings as a strategic guide for future work in embodied MLLM post-training and agent design.

%% file: Sec/appendix.tex
\clearpage
\appendix

\begin{center}
{\Large\bf}

\vspace{0.3cm}
{\large \textbf{Appendix}}
\end{center}
\vspace{0.3cm}

In this supplementary material, we aim to provide additional details to support the main content of our paper: Section~\ref{app:dataset_and_context} provides the rationale for selecting datasets across different levels of instruction granularity and defines the experimental scope and context.
Section~\ref{app:semantic_generation} details the process for generating semantic annotations for the environment. Section~\ref{app:visual_construction} presents the construction and rationale of agent-centric visual observations, supported by ablation studies and empirical evidence demonstrating the representativeness of the resulting benchmark. Section~\ref{app:prompt_design} explains the detailed workflow, action parsing, and complete prompt structure used for all agent variants. Section~\ref{app:error_analysis} introduce the custom tool for trajectory visualization and analysis, provides a detailed quantitative analysis of agent failures and successes, and illustrates common agent behaviors through several qualitative case studies. Additionally, Section~\ref{app:llm_declaration} includes our declaration on the use of large language models to aid in polishing the manuscript.

\section{Dataset Selection and Experimental Scope}
\label{app:dataset_and_context}

\subsection{Selection Rationale: The Granularity Spectrum}
Our primary criterion for dataset selection is to cover the full spectrum of linguistic granularity. To evaluate MLLM spatial reasoning comprehensively, we categorize the instruction $\mathcal{W}$ into three distinct levels of granularity. We select one representative dataset for each level to ensure broad coverage:

\begin{enumerate}
    \item \textbf{Fine-grained VLN:} $\mathcal{W}$ describes the complete sequence of actions and observations $ \{s_0, a_0, s_1, a_1, \ldots, s_T, a_T\} $ step-by-step.
    \begin{itemize}
        \item \textit{Selected Dataset:} \textbf{R2R}~\cite{anderson2018r2r}. This represents the standard fine-grained task with 22k human-annotated instructions (avg. 32 words) guiding agents along ground-truth paths of approx. 7 steps (10 meters).
    \end{itemize}
    
    \item \textbf{Coarse-grained VLN:} $\mathcal{W}$ refers to a remote, out-of-sight target at $v_T$ using high-level linguistic descriptions (\textit{e.g.}, ``the cold tap in the first bedroom on level two'').
    \begin{itemize}
        \item \textit{Selected Dataset:} \textbf{REVERIE}~\cite{qi2020reverie}. This represents the coarse-grained task. It inherits the R2R topology but uses high-level instructions (avg. 21 words) for paths ranging from 4 to 7 steps.
    \end{itemize}
    
    \item \textbf{Zero-grained VLN:} $\mathcal{W}$ consists of a single term indicating the target (\textit{e.g.}, an object category), requiring the agent to infer the path without linguistic guidance.
    \begin{itemize}
        \item \textit{Selected Dataset:} \textbf{objnav-MP3D}~\cite{batra2020objectnav}. We utilize the standard validation split of 11 scenes from the Habitat objnav dataset~\cite{savva2019habitat} in MP3D~\cite{chang2017matterport3d} to represent zero-grained navigation across 21 goal categories.
    \end{itemize}
\end{enumerate}

\subsection{Scope and Extensibility}
To isolate the reasoning capabilities of MLLMs, we strictly define the scope of this work while highlighting the extensibility of our approach:

\noindent\textbf{Focus on Spatial Reasoning:} We operate within discrete action spaces (connectivity graphs) and exclude continuous control (VLN-CE~\cite{krantz2020beyond}) or physical robot deployment. Furthermore, while REVERIE typically involves object localization, we restrict our evaluation to the navigation success only. This allows us to focus purely on sequential decision-making, spatial reasoning, and understanding action consequences, avoiding the confounding factors of sim-to-real gaps or object detection failures.

\noindent\textbf{Extensibility to Other Tasks:} In this work, we prioritize static instructions and exclude interactive datasets (e.g., CVDN~\cite{thomason2020cvdn}) or multilingual variants (e.g., RxR~\cite{anderson2020rxr}). However, our framework is designed to be task-agnostic. Future extensions to interactive or multilingual tasks can be seamlessly embedded into our framework without requiring significant architectural changes.

\section{Generation of Semantic Annotations}
\label{app:semantic_generation}

To enrich the agent's environmental understanding in our simulator-free setup, we generated two types of semantic annotations: descriptive captions for navigable markers and concise summaries for each viewpoint.

\subsection{Marker Caption Generation}
\begin{figure}[h!]
\centering
  \includegraphics[width=\columnwidth]{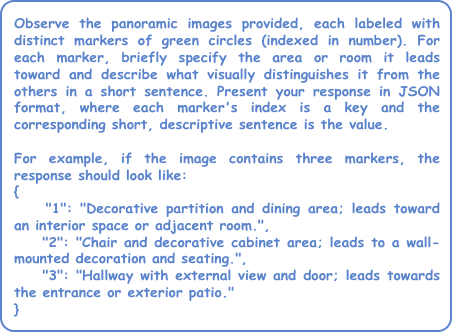}
  \caption{GPT5 prompt for generating marker caption}
  \label{fig:marker caption}
\end{figure}

The visual markers indicating navigable viewpoints in the panoramic images were annotated with short, descriptive captions. This process provides the agent with crucial semantic cues about the direction of potential paths. We used GPT-5 for this task. For each viewpoint, the model was provided with the panoramic image containing numbered visual markers and prompted to generate a JSON object mapping each marker index to a descriptive sentence. The prompt used for this captioning process is shown in Figure~\ref{fig:marker caption}.

\subsection{Viewpoint Summary Generation}
In addition to marker captions, a single, holistic summary of the scene was generated for each viewpoint to give map-based agents a global understanding of their current location. For this process, we adopt the same methodology presented in NavGPT~\citep{zhou2024navgpt}.

The generation follows a two-stage process. First, initial descriptions are generated for images from a viewpoint using the BLIP-2 model. To elicit descriptions that are rich in object details and relevant to indoor scenes, each image is fed to BLIP-2 with the prompt: ``\textit{This is a scene of}".

\begin{figure}[h!]
\centering
  \includegraphics[width=0.8\columnwidth]{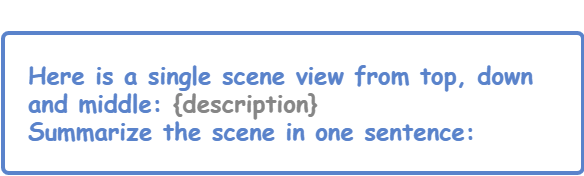}
  \caption{GPT-5 prompt for generating viewpoint summarization}
  \label{fig:summarizer}
\end{figure}

As this initial step often produces redundant information across different images of the same viewpoint, a second summarization step is employed. The descriptions generated by BLIP-2~\citep{li2023blip2} are consolidated into a single, coherent sentence using a GPT-5 summarizer. The model is prompted with the following template in Figure ~\ref{fig:summarizer}, where ``\{description\}" is replaced by the text from BLIP-2.

\begin{figure*}[t]
	\centering
    \includegraphics[width=\textwidth]{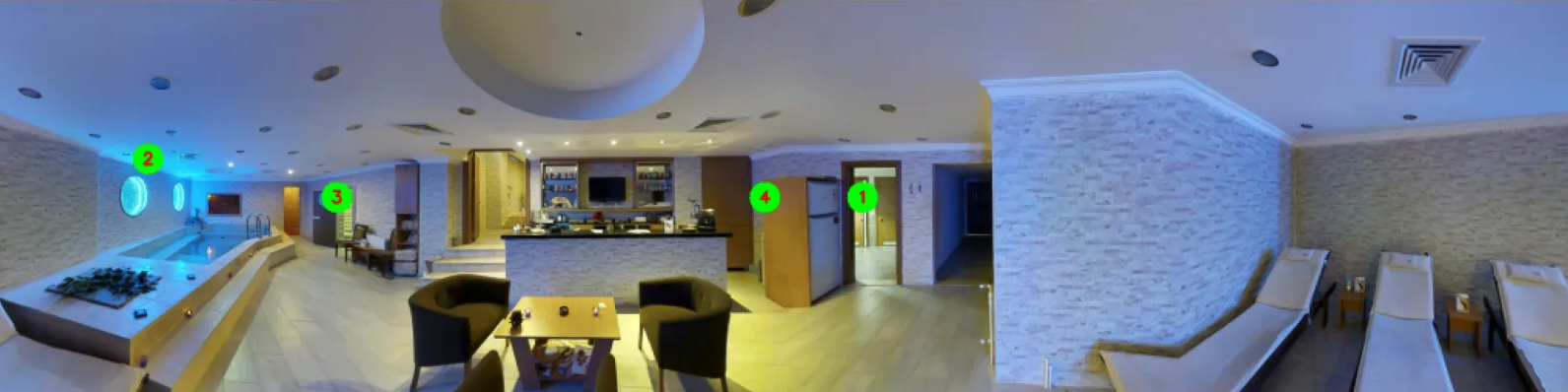}
	\caption{An example of the composite visual observation provided to the MLLM. The four pre-rendered images are stitched together in an agent-centric order (Left, Front, Right, Back) based on the agent's corrected heading. }
	\label{fig:observation_example}
\end{figure*}

\section{Construction of Agent-Centric Visual Observations}
\label{app:visual_construction}

\subsection{Panoramic Construction Details}
To provide the MLLM with a full 360-degree visual context from the agent's perspective, we construct a single panoramic image at each step. This process leverages the four pre-rendered, world-oriented images associated with each viewpoint and reorients them based on the agent's current heading. This method serves as a lightweight, simulator-free proxy for real-time rendering.

\paragraph{Pre-rendered Image Set} As described in the main paper, each viewpoint in the environment is associated with four high-resolution images, each with a 90-degree vertical Field of View (vFOV). These images are centered on the four cardinal directions relative to the global coordinate system: 0\textdegree~(North), 90\textdegree~(East), 180\textdegree~(South), and 270\textdegree~(West).

\paragraph{Heading Correction and Image Selection} Since an agent's heading is continuous (e.g., 60\textdegree), it will not always align perfectly with one of the four pre-rendered directions. To resolve this, we implement a heading correction mechanism. The agent's current continuous heading is first mapped to the closest cardinal direction. This is achieved by quantizing the heading angle to the center of the 90-degree quadrant it falls within. For instance, any agent heading $h \in [45\textdegree, 135\textdegree)$ is mapped to the 90\textdegree~image, which then serves as the agent's \textbf{Front} view.

\paragraph{Panoramic Image Composition} Once the \textbf{Front} image is determined through heading correction, the remaining three images are assigned to the agent's relative directions: \textbf{Left}, \textbf{Right}, and \textbf{Back}. These four images are then concatenated horizontally in the following order to form a single panoramic strip: [\textbf{Left}, \textbf{Front}, \textbf{Right}, \textbf{Back}].

To ensure the MLLM can correctly interpret this composite view, we explicitly annotate the image by overlaying the corresponding directional labels above each of the four segments, as illustrated in Figure~\ref{fig:observation_example}. This provides a clear, agent-centric visual input that grounds the model in its current orientation.

\subsection{Justification of Observation Format}
The choice of using a 4-image panoramic observation was not an arbitrary simplification, but a design decision informed by systematic experimentation across four observation formats. To determine the optimal visual input for the MLLM, we evaluated:
\begin{enumerate}
    \item The standard 36-image panoramic sweep (dense sampling).
    \item A 24-image variant tested in prior work~\citep{zhou2024navgpt}.
    \item A single-image stitched panorama (warped projection).
    \item Our proposed 4-image 360\textdegree~view (90\textdegree~vFOV each).
\end{enumerate}

\begin{table*}[t]
\centering
\small
\caption{\textbf{Ablation Study on Visual Observation Formats.} We compare our proposed 4-image view against standard dense sweeps (36 and 24 images) and a single stitched panorama. The 4-image format yields the highest Success Rate (SR) and Oracle Success Rate (OSR) across both MLLMs. \textbf{Bold} indicates the best performance.}
\label{tab:obs_ablation}
\resizebox{\textwidth}{!}{%
\begin{tabular}{llcccccccc}
\toprule
\textbf{Model} & \textbf{Obs. Format} & \textbf{TL} $\downarrow$ & \textbf{NE} $\downarrow$ & \textbf{SR} $\uparrow$ & \textbf{OSR} $\uparrow$ & \textbf{SPL} $\uparrow$ & \textbf{nDTW} $\uparrow$ & \textbf{SDTW} $\uparrow$ & \textbf{CLS} $\uparrow$ \\
\midrule
\multirow{4}{*}{InternVL3-8B} 
 & 36 images & 12.40 & 8.20 & 22.80 & 44.00 & 9.10 & 22.00 & 11.10 & 22.50 \\
 & 24 images & 12.10 & 7.95 & 24.50 & 46.80 & 10.40 & 23.40 & 12.00 & 24.30 \\
 & Panoramic (single) & 13.20 & 8.80 & 20.40 & 41.50 & 8.20 & 20.00 & 10.10 & 20.70 \\
 & \textbf{4-image (Ours)} & \textbf{11.74} & \textbf{7.55} & \textbf{28.00} & \textbf{50.50} & \textbf{12.61} & \textbf{25.28} & \textbf{13.38} & \textbf{26.22} \\
\midrule
\multirow{4}{*}{Qwen2.5-VL-7B} 
 & 36 images & 9.40 & 7.60 & 24.00 & 39.50 & 14.80 & 33.10 & 17.20 & 31.80 \\
 & 24 images & 9.05 & 7.30 & 25.80 & 41.70 & 15.90 & 34.40 & 17.90 & 33.20 \\
 & Panoramic (single) & 10.20 & 8.55 & 21.40 & 37.90 & 13.00 & 30.00 & 15.10 & 28.10 \\
 & \textbf{4-image (Ours)} & \textbf{8.54} & \textbf{6.99} & \textbf{27.50} & \textbf{44.00} & \textbf{17.11} & \textbf{35.97} & \textbf{18.88} & \textbf{34.85} \\
\bottomrule
\end{tabular}%
}
\end{table*}

To ensure the comparison was meaningful, we conducted these tests on two representative MLLMs (InternVL3-8B and Qwen2.5-VL-7B) on the fine-grained navigation task. As shown in Table~\ref{tab:obs_ablation}, our proposed 4-image format consistently outperforms significantly denser representations (24 or 36 images) and the single stitched panorama. 

We hypothesize that while 24 or 36 images provide more granular visual information, they significantly increase the token count and complexity, potentially overwhelming the MLLM's context window or introducing irrelevant visual noise. Conversely, the single stitched panorama likely suffers from projection distortion or loss of fine-grained detail. The 4-image format strikes an optimal balance, providing high-resolution, undistorted views of the cardinal directions without excessive token overhead.

\subsection{Benchmark Representativeness and Baseline Comparison}

\begin{figure}[h!]
\centering
\includegraphics[width=\linewidth]{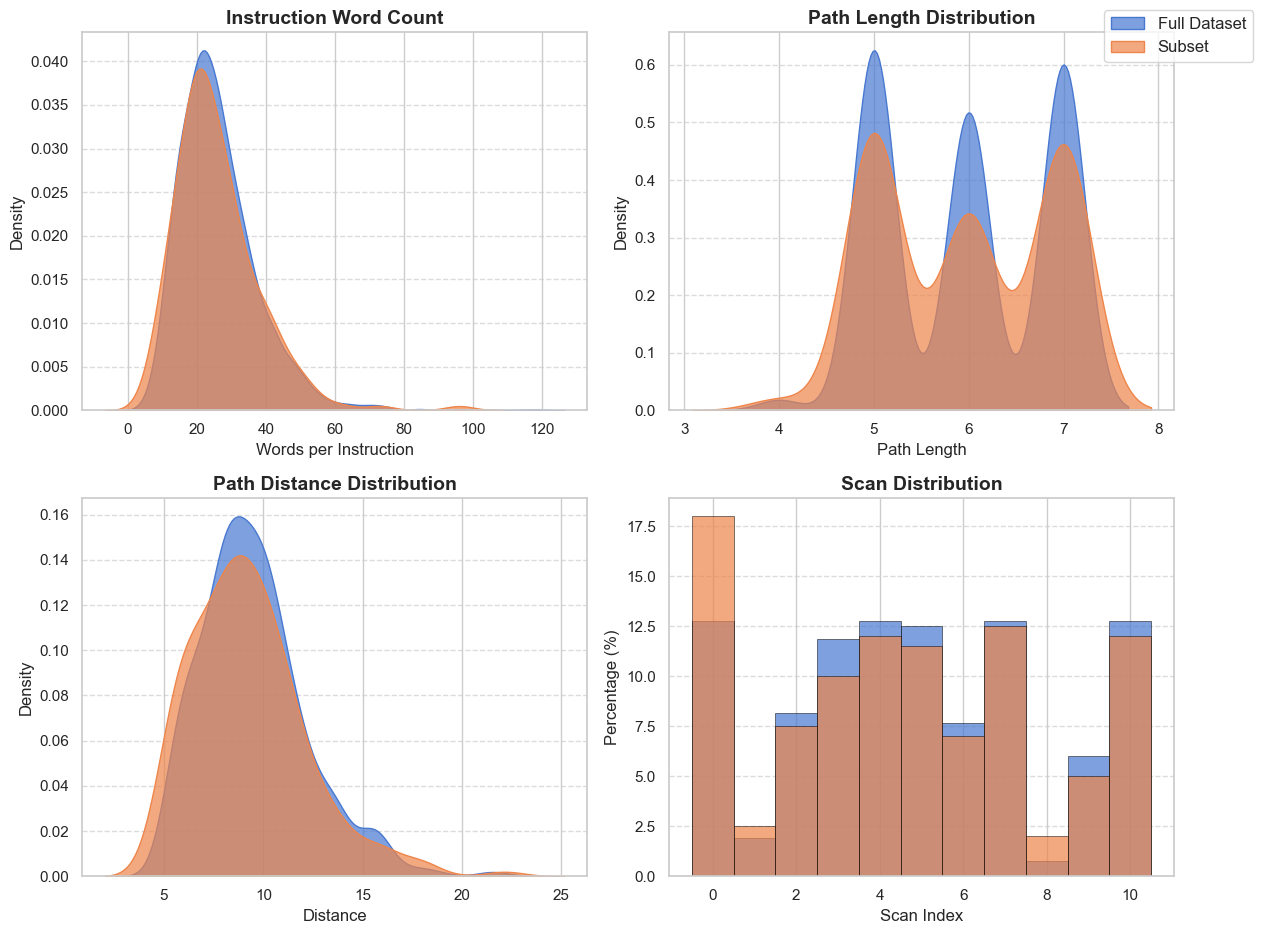}
\caption{A statistical comparison of our benchmark's fine-grained task subset data against the original R2R val\_unseen split.}
\label{fig:distribution}
\end{figure}

\input{Tab/VLM_compare}

To verify the representativeness of our dataset, we first examine the instruction and path characteristics. As illustrated in Figure~\ref{fig:distribution}, the data distribution of our sampled benchmark aligns closely with the original R2R \texttt{val\_unseen} split, indicating that our subset successfully preserves the intrinsic diversity and complexity of the full dataset. 

Complementing this statistical alignment, we evaluated a diverse set of prior methods, ranging from traditional VLN specialists (RecVLNBert~\cite{hong2020recurrent}, HAMT~\cite{chen2021hamt}, DUET~\cite{chen2022duet}) to finetuned MLLMs (NaviLLM~\cite{zheng2023towards}, NavGPT2~\cite{zhou2025navgpt2}), on both the full \texttt{val\_unseen} splits and our curated benchmark for R2R and REVERIE. The results reveal a strong performance correlation: key metrics such as Success Rate (SR) and Success weighted by Path Length (SPL) on our benchmark closely track the performance on the full splits, with deviations typically within a 2 to 3 percentage point margin. This close alignment confirms that our stratified sampling successfully captures the intrinsic difficulty and diversity of the original datasets, establishing our benchmark as a reliable proxy for full-scale evaluation. 

When comparing these baselines to our proposed framework, however, a significant performance gap remains. As referenced in Table~\ref{tab:compare_vlm}, VLN specialists and finetuned agents achieve substantially higher success rates (e.g., 72\% SR on R2R) compared to the zero-shot average. Despite this gap, our zero-shot agents demonstrate non-trivial navigation capabilities, establishing a crucial baseline for generalization without the cost of task-specific training.

% ---- Prompt part
\section{Agent Workflow and Prompt Design}
\label{app:prompt_design}

A central component of our framework lies in the design of agent and prompts that guide multimodal large language models (MLLMs) to behave as navigation agents. Since the main paper provides only a brief overview, we expand here with a full account of the workflow, structure, components, and variations used across all eight agent types. Our agents are divided into two families: \textit{Text Summarization as Memory (NavGPT)}, which relies purely on local observations and history, and \textit{Text Map as Memory (MapGPT)}, which augments navigation with dynamically constructed topological maps. Within each family, we instantiate four variants: a baseline version, a chain-of-thought (CoT) agent, a reflection-enabled agent, and a combined CoT+Reflection agent. This section explains the design philosophy of each family and the detailed structure of their prompts.

\subsection{Overall Agent Workflow}
\label{app:agent_workflow}

Although our agents are implemented as executable modules, their behavior is fundamentally driven by prompt design. Conceptually, each agent operates in a closed interaction loop with a self-defined simulator wrapper, where perception, reasoning, and action selection are mediated entirely through structured prompting.

At each navigation step, the agent first interacts with our self-designed simulator via the runner module to retrieve the full set of environmental information and visual observations required for decision making. This includes the panoramic visual input, the navigation instruction, the agent’s current heading and elevation, the navigation history, and the set of navigable action options. For text map memory agents, this additionally includes the dynamically constructed topological map and node connectivity information described in Sections~\ref{app:text-summary} and~\ref{app:text-map}.

These elements are then formatted into a prompt following the unified structure introduced in Section~\ref{app:overall_prompt}. The system component provides persistent role definitions, navigation rules, and output constraints, while the task component injects step-specific state information. For multimodal models, the prompt is paired with the corresponding visual observation and passed to the MLLM as a single inference call.

Upon receiving the prompt, the MLLM generates a structured textual output that encodes its navigation decision. Depending on the agent variant, this output may include explicit reasoning traces (CoT), reflective self-evaluation, or both, as detailed in Sections~\ref{app:text-summary}–\ref{app:cot_reasoning_example}. Crucially, regardless of internal reasoning style, the output always conforms to a strict, predefined format that exposes a single actionable decision.

The agent then parses this output to extract the intended action and feeds it back to the simulator to execute the corresponding movement. The environment transitions to a new state, and the interaction loop repeats until the agent selects the stop action or reaches the maximum step limit. This design ensures that all agent behavior—state interpretation, reasoning, and control—is governed by prompt construction and output parsing, rather than task-specific procedural logic.

The detailed mechanics of prompt composition are described in Sections~\ref{app:overall_prompt}–\ref{app:cot_reasoning_example}, while the rules and implementation of action extraction from model outputs are specified in Section~\ref{app:parse_action}.

\subsection{Overall Prompt Structure}
\label{app:overall_prompt}
\begin{figure}[h!]
\centering
  \includegraphics[width=\columnwidth]{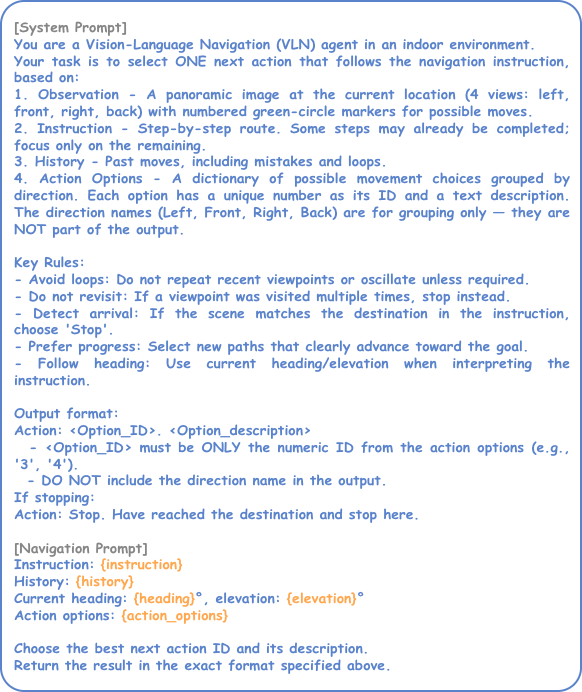}
  \caption{Text summarization memory baseline agent prompt structure}
  \label{fig:prompt_structure}
\end{figure}

All prompts are composed of two distinct parts: the \textit{system} and the \textit{task} component. The system portion defines the global context of the agent, introducing the VLN setting, enumerating the input elements, and stating the rules the model must follow when reasoning about navigation. It also enforces the strict output format required for downstream evaluation. The task portion is dynamic, providing the specific input to the agent at each time step: the instruction, navigation history, agent orientation, and the set of navigable options. Together, these two components establish both the constraints and the situational awareness necessary for coherent decision making. Figure~\ref{fig:prompt_structure} illustrates an example of the full text summarization as memory baseline prompt.

\subsection{Text Summarization as Memory Agents}
\label{app:text-summary}
The NavGPT-style agents are designed to operate with information that would be available in a simulator-based VLN setup but translated into our simulator-free representation. The system prompt explicitly instructs the model to select a single next action, referencing only the option identifiers, and to obey a series of rules that reduce common navigation errors such as looping, oscillation, and premature stopping. 

The task inputs are carefully structured. The navigation \textbf{instruction} is passed in verbatim, ensuring the model has access to the original language guidance. The \textbf{history} describes prior movements in natural language, with each step recorded as a turning angle, forward displacement, and the semantic description of the destination viewpoint. This representation provides both spatial reasoning cues and semantic grounding. The agent’s \textbf{current heading and elevation} are provided as numerical values, anchoring the model’s interpretation of orientation. Finally, the \textbf{action options} are represented as a dictionary keyed by relative directions, where each entry contains a marker ID and a semantic description of the corresponding navigable viewpoint, along with an explicit “Stop” action. This structured but naturalistic representation ensures the model can ground its decisions in both geometry and semantics.

\subsection{Reasoning-Enhanced Text Summarization Memory Agent Variants}
To probe the role of explicit reasoning, we introduce three reasoning-augmented variants of text summarization memory agent. In the CoT version, the system prompt is modified so that the agent first produces a reasoning trace encapsulated in \verb|<Reasoning>| tags before committing to its final action choice. This design encourages more transparent step-by-step deliberation. The Reflection variant modifies the output format further: after producing an action, the agent generates a reflective evaluation wrapped in \verb|<Reflection>| tags, followed by a \verb|<Final Decision>| statement declaring whether to keep or revise its action. If the reflection deems the decision unsound, the agent replans rather than moving. The CoT+Reflection version combines both mechanisms, first reasoning explicitly and then reflecting on the proposed choice, providing the richest form of introspective navigation. These modifications shift the model from direct action prediction toward a more deliberative, self-monitoring behavior.

\begin{figure}[htbp]
\centering
  \includegraphics[width=\columnwidth]{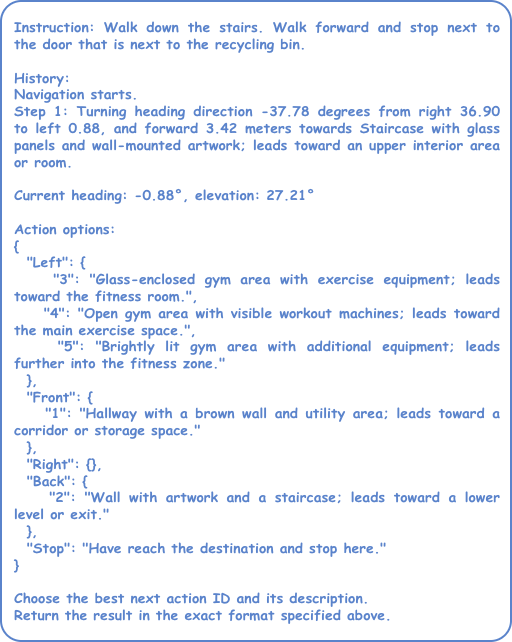}
  \caption{Text map memory agent prompt example (Step 1)}
  \label{fig:mapgpt_prompt}
\end{figure}

\subsection{Text Map as Memory Agents}
\label{app:text-map}
While NavGPT-style focuses on local decision making, the MapGPT-style agents introduces a form of spatial memory through a dynamically constructed topological graph. At each step, the agent augments its prompt with a \textbf{map connectivity} field, expressed in natural language, that lists adjacency relationships between viewpoints (e.g., ``node\_0 is connected to node\_1, node\_2''). This evolving graph representation enables the MLLM to reason not only about immediate action choices but also about the broader connectivity of the explored environment.

The navigation history for text map memory agents is likewise enriched. Instead of recording only motion trajectories, it includes the current node identifier, a semantic description of the viewpoint, and the sets of visited and unvisited nodes. This structure gives the agent both a local semantic grounding and a global perspective on the exploration state. A complete prompt example after one navigation step is illustrated in Figure~\ref{fig:mapgpt_prompt}.

\subsection{Reasoning-Enhanced Text Map Memory Agent Variants}
The CoT, Reflection, and CoT+Reflection augmentations are applied to text map memory agents in the same manner as for text summarization agents, modifying only the output structure while retaining the additional map input. Thus, the text map memory agents explores how explicit reasoning interacts not just with semantic cues, but also with global topological memory. 

\subsection{Example CoT Reasoning}
\label{app:cot_reasoning_example}
\label{app:example cot}
\begin{figure}[h!]
\centering
\includegraphics[width=\linewidth]{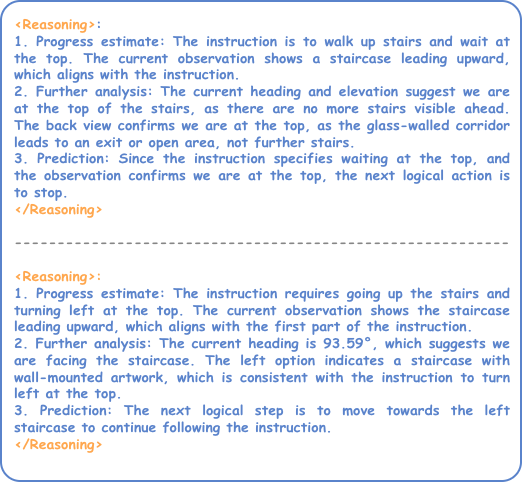}
\caption{Two examples of chain-of-thought (CoT) reasoning generated by Qwen2.5-VL-7B. Both cases demonstrate structured step-wise reasoning but limited integration of historical context.}
\label{fig:cot_examples}
\end{figure}

Figure~\ref{fig:cot_examples} presents two representative CoT outputs. In both cases, the model correctly decomposes the navigation instruction into progress estimation, further analysis, and prediction. However, despite being provided with full interaction history, the reasoning predominantly relies on local observations and the most recent instruction, while largely neglecting prior context.

This illustrates a broader issue: even when the input tokens are well within the context window, the model exhibits poor reasoning fidelity across multiple rounds of context. Instead of leveraging accumulated history for richer reasoning, the MLLM tends to perform single-turn grounding of the immediate observation. This behavior highlights a critical limitation for VLN tasks, where successful navigation often depends on integrating long-term history with dynamic, stepwise decision-making.

\subsection{Details of Action Parsing}
\label{app:parse_action}

After the MLLM generates a structured textual response, the agent must convert this free-form output into an executable navigation action. Although the prompting enforces a strict output format, in practice MLLM outputs may contain formatting noise, partial deviations, or additional explanatory text. We therefore adopt a robust, rule-based action parsing strategy that differs slightly across agent variants.

\paragraph{Baseline and CoT Agents.}
For the baseline and chain-of-thought (CoT) agents, the parsing procedure is intentionally minimal. Since these agents are instructed to output only a single action declaration, the parser searches for an \texttt{Action} field and extracts the associated action value, which may be expressed either as a numeric viewpoint identifier or as a relative direction token (e.g., \texttt{Left}, \texttt{Front}). No additional fields are expected, and any successfully extracted action is immediately validated against the current set of navigable viewpoints before being executed.

\paragraph{Reflection-Enabled Agents.}
Reflection-based agents require more elaborate parsing, as their outputs may contain multiple structured components, including reflective analysis and decision revision. Specifically, these agents may produce optional \texttt{Reflection} and \texttt{Decision} (or \texttt{Final Decision}) fields in addition to the final \texttt{Action}.

To accommodate this variability, we employ a two-stage parsing strategy. First, the parser robustly extracts the action value by searching for an \texttt{Action} declaration with flexible tolerance to formatting noise, allowing for extra characters, markdown symbols, or minor deviations around the action delimiter. The parser prioritizes numeric viewpoint identifiers when present, and falls back to direction-based actions otherwise. To reduce false positives, the search is constrained to a local window following the detected action marker.

Second, if present, the parser extracts the optional \texttt{Reflection} and \texttt{Decision} segments using separate pattern matching rules. These fields are treated as auxiliary metadata: they do not influence control flow directly but are preserved for logging, analysis, and qualitative inspection of the agent’s self-evaluation behavior.

\paragraph{Action Validation and Execution.}
Regardless of agent type, all parsed actions undergo a strict validation step. Numeric actions are checked against the current viewpoint-to-node mapping and verified to correspond to a navigable location in the simulator. Direction-based actions are resolved into concrete viewpoint identifiers via the provided action option dictionary and validated in the same manner. If an extracted action does not correspond to a valid navigable transition, the agent raises a parsing exception and re-prompts the model with an explicit error signal.

In addition to movement actions, the parser also detects explicit termination signals (i.e., the stop action) and converts them into a terminal agent state. Parsing failures occur only when no valid \texttt{Action} field can be identified; in such cases, the output is treated as malformed and returned to the MLLM for correction.

Overall, regardless of reasoning verbosity, this parsing design ensures that all agent variants produce a single, well-defined navigation action at each step, while remaining robust to minor deviations in model-generated text. Importantly, the parser enforces a clear separation between reasoning content and control decisions, preserving the prompt-centric nature of the agent framework.

% Episode visualizer
\section{Episode-level Analysis}
\label{app:error_analysis}

\subsection{Trajectory Visualization and Analysis Tool}

\begin{figure*}[h!]
    \centering
    \includegraphics[width=0.85\textwidth]{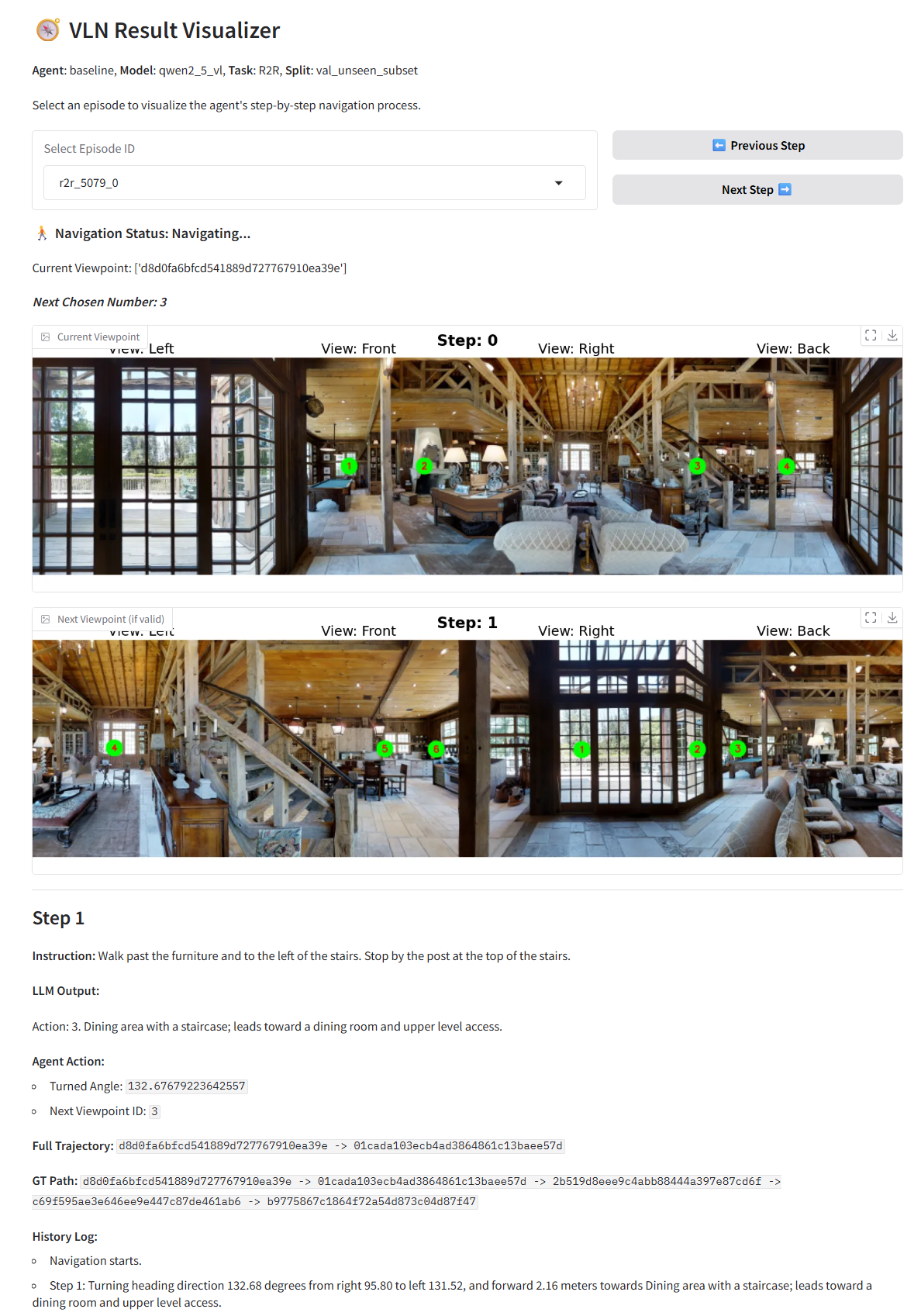}
    \caption{The main interface of the VLN Result Visualizer.}
    \label{fig:visualizer_main}
\end{figure*}

This section details the custom tool developed for the qualitative analysis of agent trajectories. The tool, named the VLN Result Visualizer, developed using Gradio, provides an interactive interface for a step-by-step inspection of any navigation episode, which is crucial for understanding the nuances of agent behavior beyond aggregate metrics.

The visualizer is built entirely using the Gradio framework. Its primary function is to parse the evaluation result files and present the information in a human-readable format. At the top of the interface, a user can specify the configuration used during evaluation, including the \textbf{agent type}, \textbf{MLLM model}, \textbf{task}, and \textbf{data split}. Once a configuration is loaded, a dropdown menu is populated with all episode IDs from that run, allowing for the selection of any specific trajectory for analysis.

The core of the interface, shown in Figure~\ref{fig:visualizer_main}, is the visual observation panel. It displays the agent's panoramic view for the Current Viewpoint and, if a valid move is made, the panoramic view of the chosen Next Viewpoint. Each panoramic image is a composite of four individual images presented in the agent-centric order of [\textbf{Left}, \textbf{Front}, \textbf{Right}, \textbf{Back}], with the global orientation (e.g., ``View: Right") explicitly labeled above each segment. Navigable options are clearly marked with green circular markers.

Below the visual panel, detailed textual information for the current step is provided. This includes: the step number, the original navigation instruction, the raw LLM Output, the parsed agent action (turn angle and forward distance), the full trajectory path taken so far, and the ground truth path. Critically, the tool also flags the exact step at which the agent's path first deviated from the ground truth, enabling quick identification of crucial mistakes. This is followed by the complete history log that was fed into the model at that step, allowing for an in-depth analysis of the agent's reasoning context.

To further enhance usability, the interface provides status indicators directly on the display during navigation. The ID of the current viewpoint and the numeric ID of the next chosen marker are displayed at the top of the screen, providing immediate context without needing to consult the text logs. 

At the conclusion of each episode, a final summary panel presents the quantitative evaluation metrics, such as Success Rate, SPL, Navigation Error etc., offering a direct link between the agent's step-by-step actions and its final performance score. This tool was indispensable for conducting the detailed error analysis presented in this paper.

% Error and success analysis
\subsection{Detailed Error and Success Analysis}

This section provides a detailed analysis of agent performance across 200 navigation trajectories. By breaking down both failures and successes, we can identify the primary challenges MLLM-based agents face in the VLN task.

\begin{figure}[htbp]
\centering
    \includegraphics[width=\columnwidth]{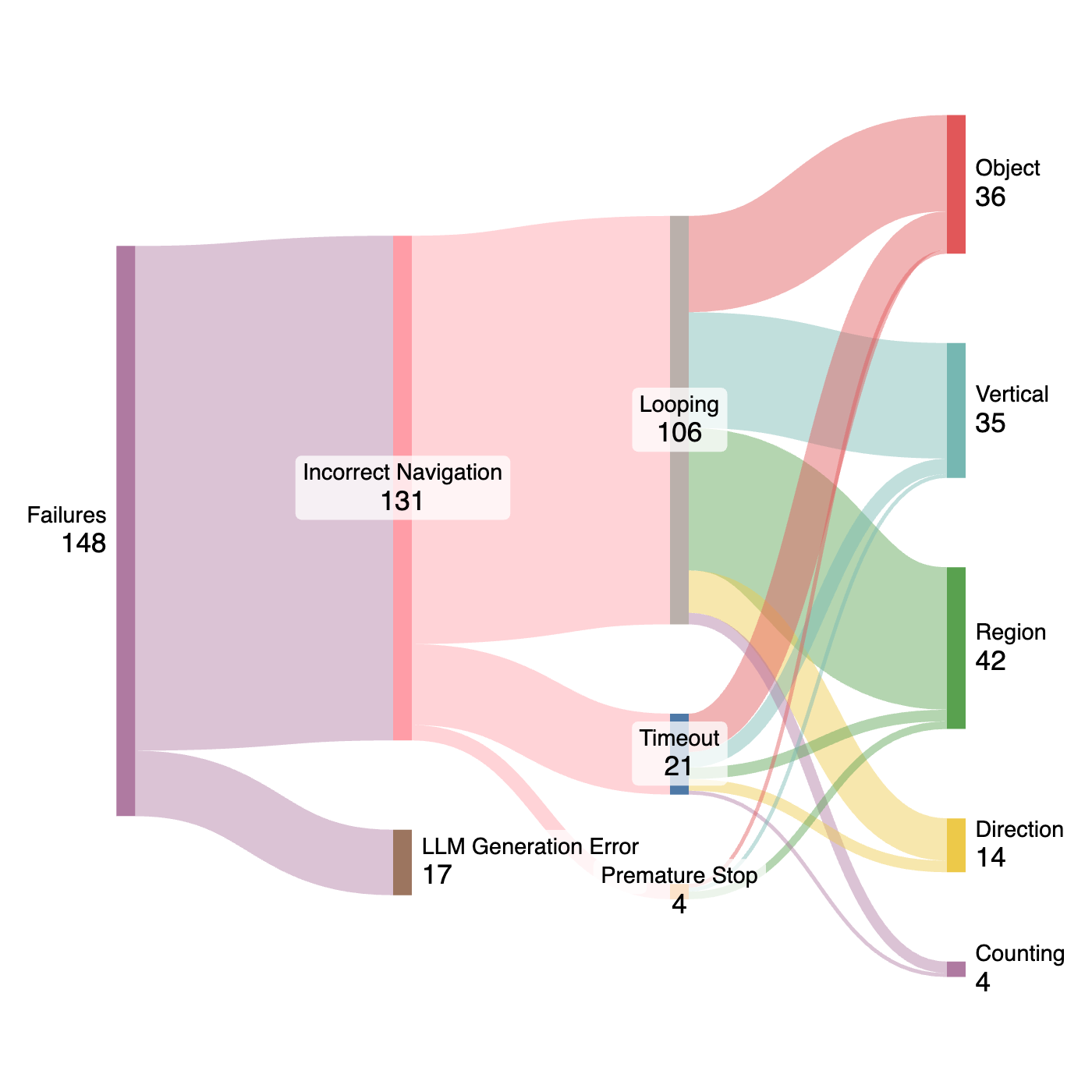}
    \caption{Analysis of error sources in 148 failure episodes.}
    \label{fig:fail}
\end{figure}

The results reveal a significant performance gap, with 148 failures compared to only 52 successes. An initial breakdown of the failures, as shown in Figure~\ref{fig:fail}, indicates that the vast majority (131 out of 148) stem from \textbf{incorrect navigation} rather than technical \textbf{MLLM Generation Errors} (17 cases). This suggests that while the models are generally capable of producing valid actions, their decision-making logic is the primary point of failure.

\begin{figure}[htbp]
\centering
    \includegraphics[width=\columnwidth]{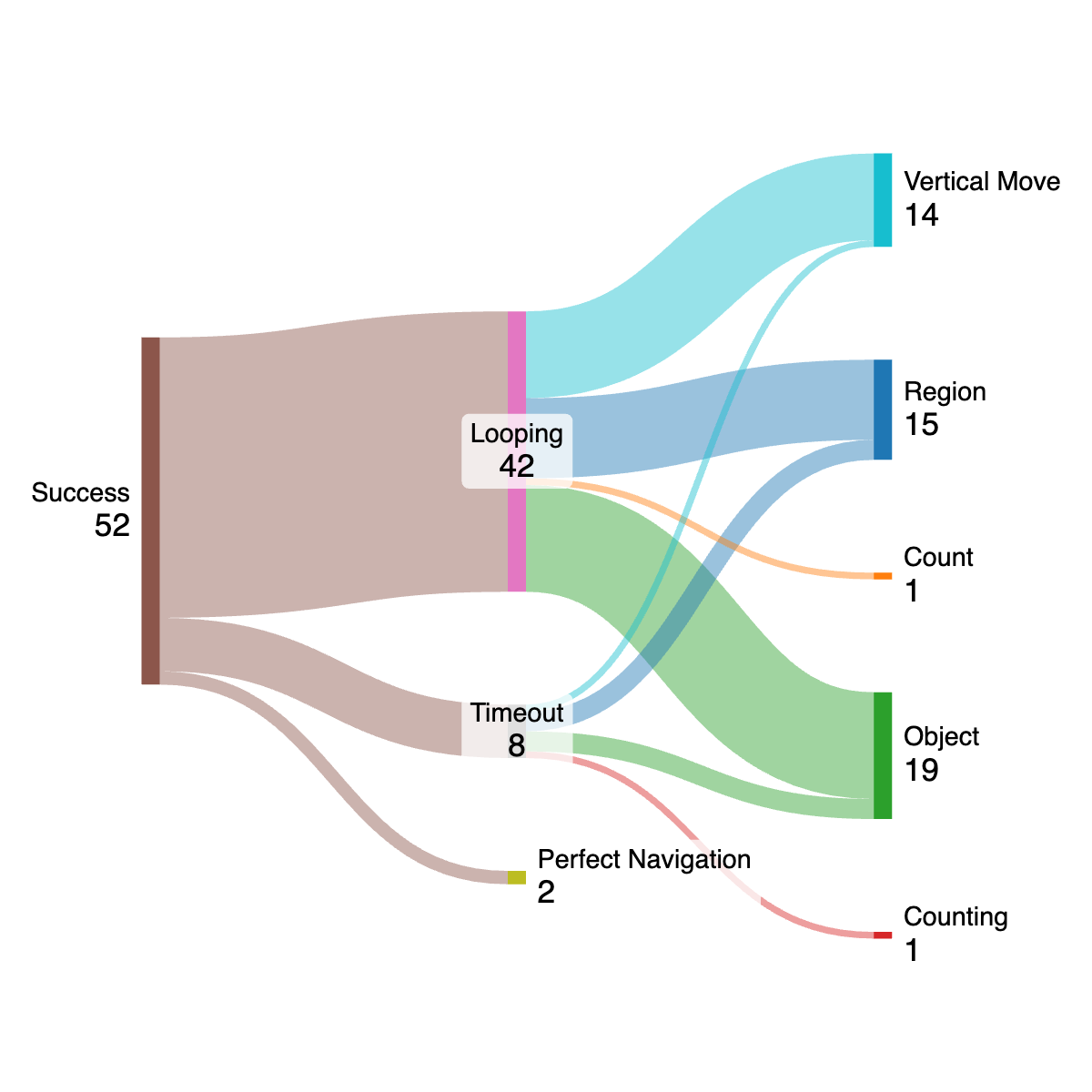}
    \caption{Analysis of navigation behavior in 52 successful episodes.}
    \label{fig:success}
\end{figure}

Within the incorrect navigation errors, \textbf{looping} is the most dominant failure mode, accounting for a remarkable 106 cases. This behavior, where the agent repeatedly revisits the same viewpoints, points to a fundamental difficulty in spatial awareness and state tracking. The root causes for these loops, as well as for timeouts, are primarily failures in high-level scene understanding. Specifically, \textbf{region recognition} (37 cases in looping), \textbf{vertical movement understanding} (30 cases), and \textbf{object detection} (25 cases) are the most frequent triggers for getting stuck. This highlights the agent's struggle to match abstract instructions (e.g., ``go to the kitchen", ``go upstairs") with visual evidence.

Conversely, an analysis of the 52 successful trajectories provides a more nuanced picture of the agent's capabilities, as illustrated in Figure~\ref{fig:success}. A striking finding is that only 2 trajectories were completed perfectly. The vast majority of successes (42 cases) were achieved despite the agent exhibiting \textbf{looping behavior}, typically near the target. This suggests that while agents can eventually recover from local confusion, their navigation is often highly inefficient. The challenges in these near-success cases mirror those in the failures: difficulties with \textbf{object recognition} (16 cases), \textbf{vertical movement} (13 cases), and \textbf{region understanding} (11 cases) still persist, causing inefficiency even when the final goal is reached.

In conclusion, the data indicates that the primary obstacle for these MLLM agents is not language generation but robust spatial and semantic reasoning. The pervasive issue of looping, both in failed and successful episodes, underscores a weakness in creating and maintaining a stable understanding of the environment. Future work should focus on enhancing these core reasoning capabilities to improve both the reliability and efficiency of navigation.

\subsection{Case Studies}
\label{app:case_study}
In this section, we provide a qualitative analysis of five navigation episodes to illustrate the agent's common behaviors, highlighting both its capabilities and frequent failure modes.

\begin{figure*}[p]
    \centering
    \includegraphics[width=\textwidth]{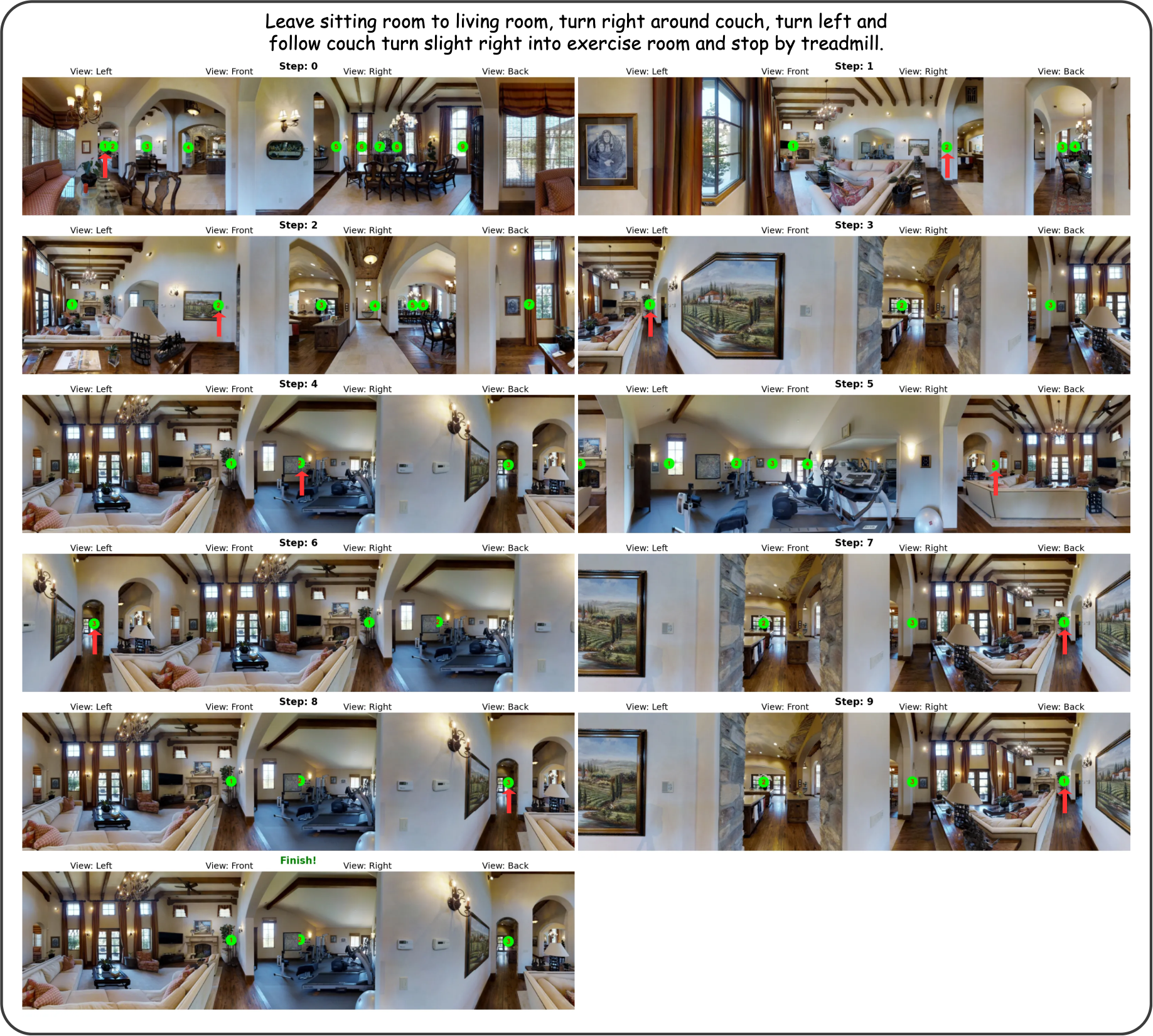}
    \caption{A successful but inefficient trajectory. After observing the target treadmill, the agent loops around the room before stopping.}
    \label{fig:788}
\end{figure*}

\begin{figure*}[p]
    \centering
    \includegraphics[width=\textwidth]{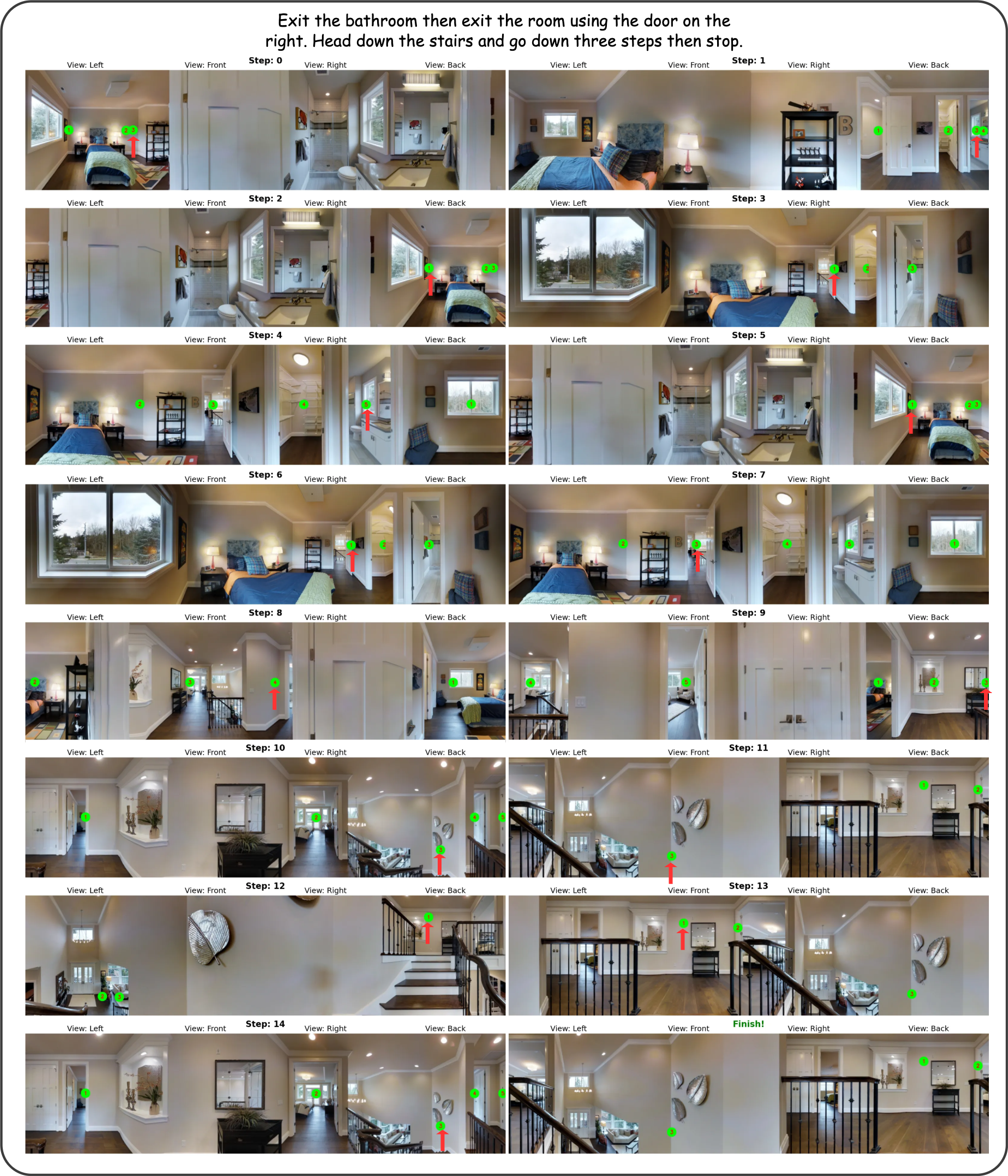}
    \caption{A successful episode showcasing recovery capabilities. However, the agent exhibits looping behavior during vertical movement on the stairs.}
    \label{fig:3272}
\end{figure*}

Figures~\ref{fig:788} and \ref{fig:3272} showcase successful episodes that also reveal subtle inefficiencies. In Figure \ref{fig:788}, the agent correctly identifies the target treadmill but exhibits redundant behavior by moving away and looping back before executing the final stop action. Similarly, the episode in Figure \ref{fig:3272} demonstrates a strong recovery capability, yet the agent struggles with precise vertical positioning, causing it to loop on the stairs rather than stopping at the correct step.

\begin{figure*}[p]
    \centering
    \includegraphics[width=\textwidth]{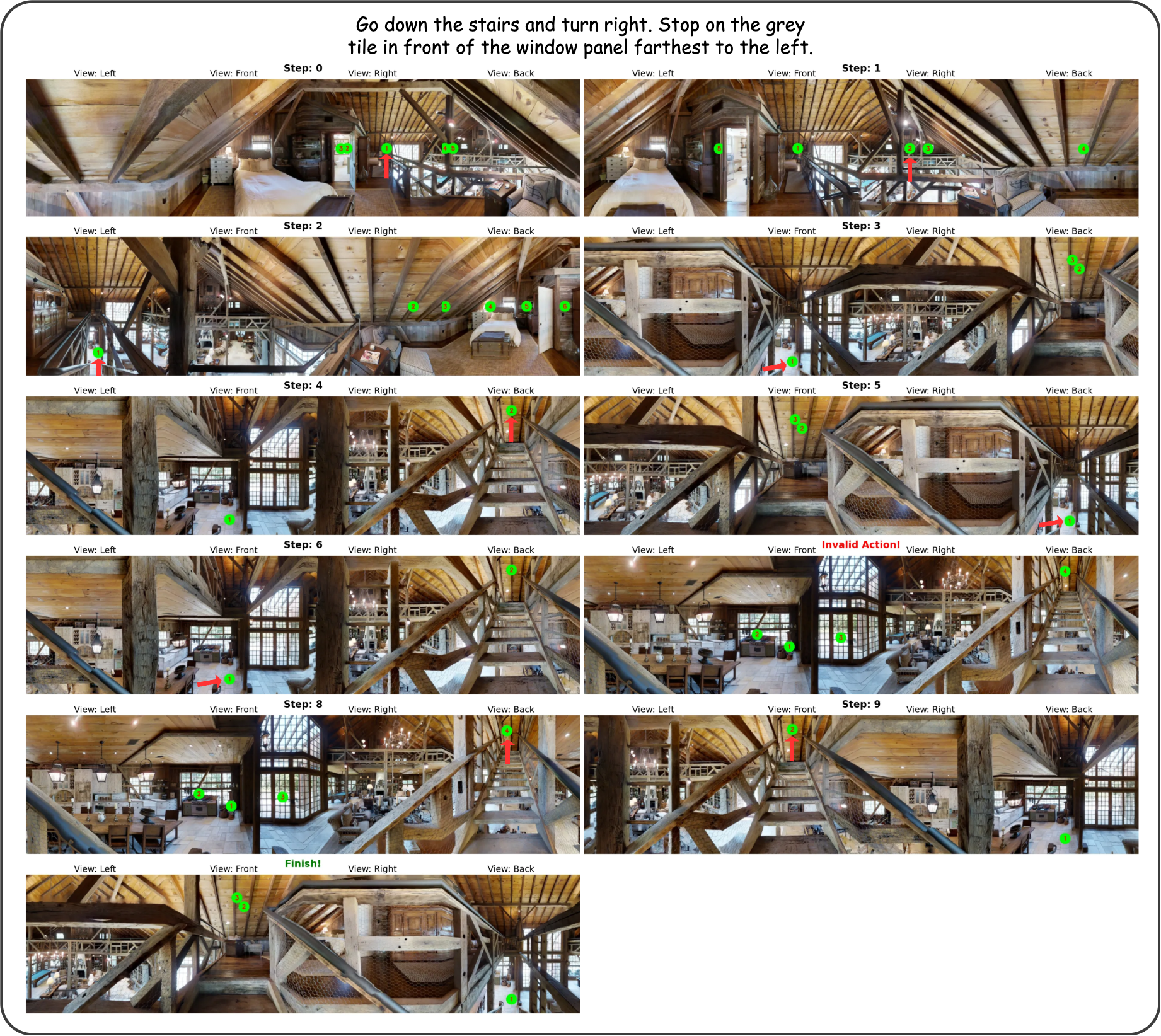}
    \caption{A failure case with oracle success. The agent reaches the correct general area but fails to stop, getting stuck in a loop on the stairs.}
    \label{fig:2194}
\end{figure*}

\begin{figure*}[p]
    \centering
    \includegraphics[width=\textwidth]{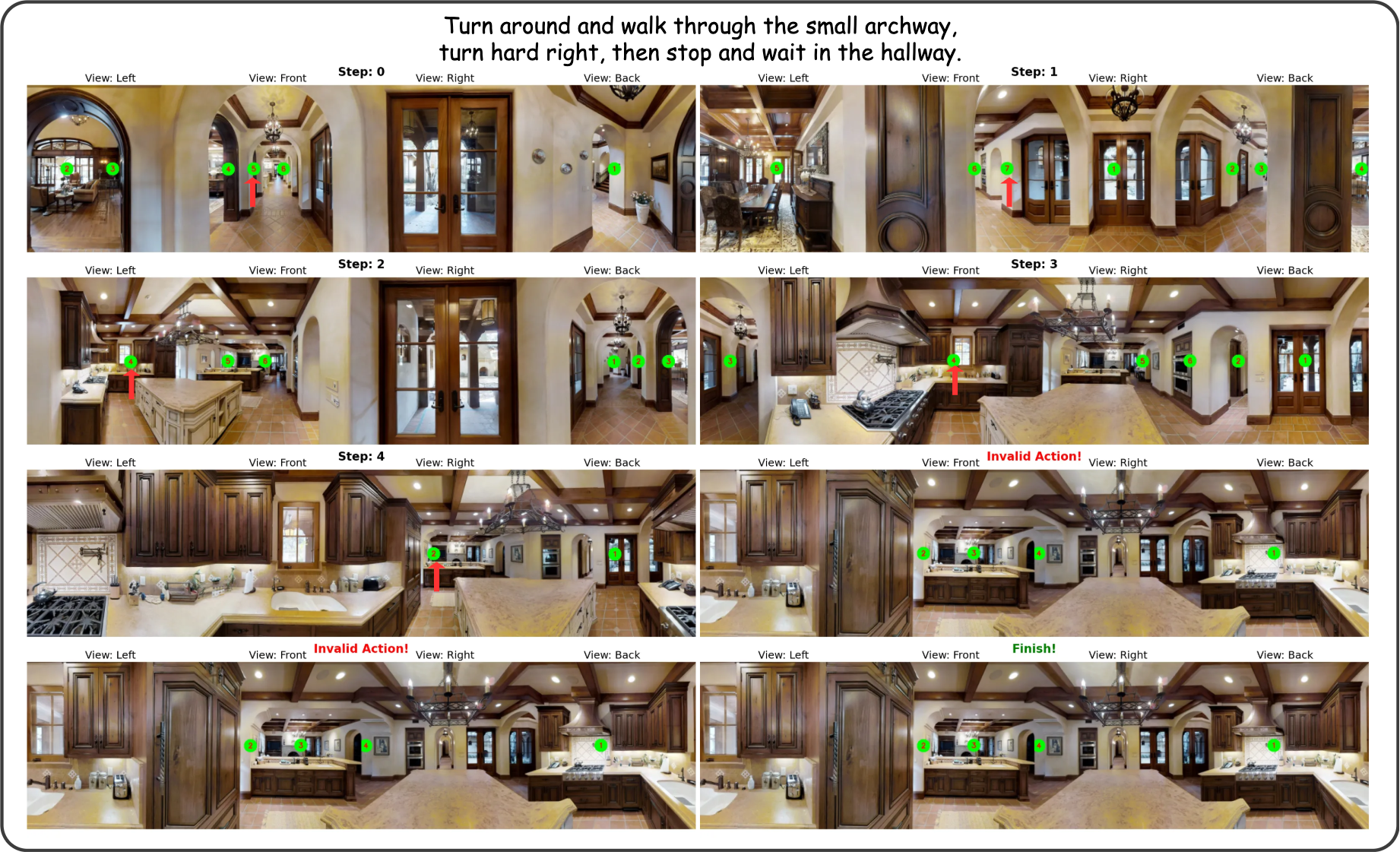}
    \caption{Navigation failure due to misinterpreting a directional instruction and a model generation error that produced an invalid action.}
    \label{fig:5203}
\end{figure*}

\begin{figure*}[p]
    \centering
    \includegraphics[width=\textwidth]{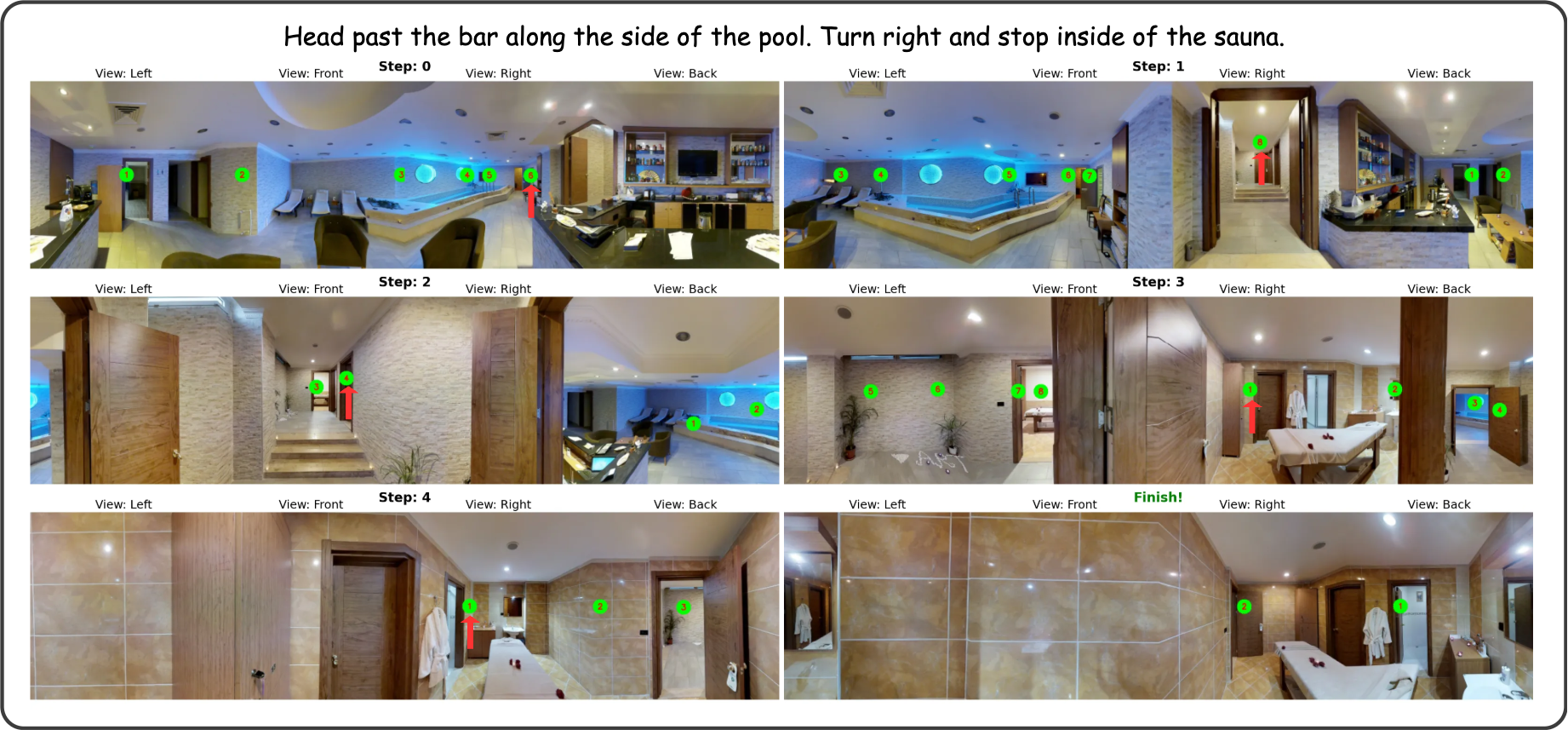}
    \caption{A failure episode caused by the agent's inability to understand and navigate into the specified target region (sauna).}
    \label{fig:2662}
\end{figure*}

Conversely, Figures \ref{fig:2194}, \ref{fig:5203}, and \ref{fig:2662} depict common failure scenarios. The trajectory in Figure \ref{fig:2194} represents a case of 'oracle success,' where the agent navigates to the immediate vicinity of the destination downstairs but ultimately fails by getting trapped in a repetitive loop on the staircase. Figure \ref{fig:5203} illustrates a multi-faceted failure; the agent not only fails to ground the directional instruction and identify the `hallway' but also produces a malformed output, resulting in an un-parsable command and an invalid action error. Finally, Figure \ref{fig:2662} demonstrates a failure in semantic region understanding, where the agent is unable to correctly interpret the goal of stopping `inside of the sauna'.

\section{The Use of Large Language Models (LLMs)}
\label{app:llm_declaration}
As disclosed, we utilized LLMs (GPT5, Google Gemini etc.) to aid in polishing the manuscript's prose. Its role was to improve grammatical correctness and sentence clarity, with all final content being reviewed and approved by the authors, who take full responsibility for this work.

%% file: Tab/VLM_compare.tex
\definecolor{tableheadgray}{gray}{0.9}

\begin{table*}[t]
\centering
\resizebox{\textwidth}{!}{
\definecolor{Gray}{gray}{0.94}
\begin{tabular}{l|ccccc|ccccc|ccccc|ccccc}
\toprule
\midrule
\multicolumn{1}{c|}{\multirow{3}{*}{Methods}} & \multicolumn{10}{c|}{R2R} & \multicolumn{10}{c}{REVERIE} \\ 
\cmidrule(lr){2-11}
\cmidrule(lr){12-21}

& \multicolumn{5}{c|}{Val Unseen} & \multicolumn{5}{c|}{Subset} & \multicolumn{5}{c|}{Val Unseen} & \multicolumn{5}{c}{Subset} \\
\cmidrule(lr){2-6}
\cmidrule(lr){7-11}
\cmidrule(lr){12-16}
\cmidrule(lr){17-21}
 & TL & NE & OSR & SR & SPL 
 & TL & NE & OSR & SR & SPL 
 & TL & NE & OSR & SR & SPL 
 & TL & NE & OSR & SR & SPL \\ 
\midrule
\midrule

\rowcolor{tableheadgray}\multicolumn{21}{l}{\emph{VLN Specialist}:}\\
VLNBERT~\citeyearpar{hong2020recurrent} & 12.01 & 3.93 & 70 & 63 & 57 & 12.06 & 3.76 & 70 & 63 & 56 & 13.98 & 4.18 & 35 & 31 & 25 & 14.07 & 4.22 & 34 & 31 & 24 \\
HAMT~\citeyearpar{chen2021hamt} & 12.14 & 3.92 & 71 & 63 & 58 & 12.10 & 3.51 & 73 & 65 & 59 & 14.08 & 4.12 & 37 & 33 & 30 & 14.62 & 4.11 & 37 & 32 & 29 \\
DUET~\citeyearpar{chen2022duet} & 13.94 & 3.31 & 81 & 72 & 60 & 13.25 & 3.54 & 79 & 71 & 60 & 22.11 & 3.93 & 51 & 47 & 34 & 22.28 & 3.96 & 49 & 43 & 32 \\

\midrule
\rowcolor{tableheadgray}\multicolumn{21}{l}{\emph{Finetuned MLLM}:}\\
NaviLLM~\citeyearpar{zheng2023towards} & 12.81 & 3.38 & 81 & 66 & 54 & 15.80 & 3.32 & 86 & 66 & 55 & 16.04 & 5.76 & 54 & 45 & 37 & 19.02 & 5.80 & 55 & 34 & 27 \\
NavGPT-2~\citeyearpar{zhou2025navgpt2} & 12.79 & 3.35 & 78 & 70 & 67 & 12.39 & 3.04 & 82 & 74 & 62 & - & - & - & - & - & - & - & - & - & - \\

% \midrule
% \rowcolor{tableheadgray}\multicolumn{21}{l}{\emph{Zero-shot MLLM}:}\\
% InternVL3-2B & 20.69 & 9.19 & 21 & 9 & 3 & 20.36 & 8.56 & 27 & 14 & 5 & 21.57 & 10.10 & 17 & 9 & 4 & 21.71 & 10.18 & 16 & 7 & 3 \\
% InternVL3-8B & 23.23 & 7.97 & 42 & 21 & 9 & 23.77 & 7.55 & 51 & 28 & 13 & 24.45 & 8.85 & 29 & 19 & 7 & 24.52 & 9.25 & 31 & 20 & 7 \\
% LLaVA-OV-7B & 16.87 & 8.22 & 22 & 12 & 6 & 16.58 & 8.40 & 21 & 12 & 5 & 21.08 & 9.09 & 22 & 14 & 6 & 20.86 & 9.35 & 20 & 15 & 5 \\
% Qwen2.5-VL & 14.89 & 7.63 & 32 & 20 & 13 & 17.20 & 6.99 & 44 & 28 & 17 & 17.59 & 8.75 & 23 & 16 & 7 & 18.28 & 8.55 & 27 & 19 & 9 \\

\bottomrule
\end{tabular}}
\caption{Performance of baseline agents on the R2R and REVERIE tasks, with results compared across the previous and our benchmark.}
\label{tab:compare_vlm}
\end{table*}